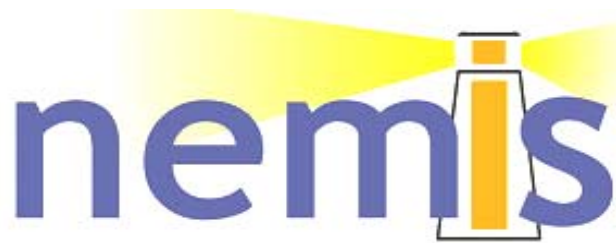

**N**etwork of
**E**xcellence in Text
**M**ining &
**I**ts applications in
**S**tatistics

**IST- 2001 - 37574**

Date: 31/08/2003

Deliverable No: D-3.1

Deliverable Title: Report of WG1
State of the Art, Evaluation and Recommendations
regarding "Document Processing and Visualization
Techniques".

Work Package: 3 (Establishment and Operation of
Working Groups)

**http://nemis.cti.gr**

Prepared By: EPFL, Switzerland

| | |
|---|---|
| **Deliverable Type:** | **R (report)** |
| **Security Status:** | **PUB (public)** |
| **Contractual Delivery Date:** | **31/08/2003** |
| **Actual Delivery Date:** | **31/08/2003** |
| **Responsible Partner:** | **EPFL** |
| **Contributing Partners:** | **–** |

**Abstract:**  Several Networks of Excellence have been set up in the framework of the European FP5 research program. Among these Networks of Excellence, the NEMIS project focuses on the field of Text Mining. Within this field, document processing and visualization was identified as one of the key topics and the WG1 working group was created in the NEMIS project, to carry out a detailed survey of techniques associated with the text mining process and to identify the relevant research topics in related research areas.  In this document we present the results of this comprehensive survey. The report includes a description of the current state-of-the-art and practice, a roadmap for follow-up research in the identified areas, and recommendations for anticipated technological development in the domain of text mining. In the part dedicated to document processing, the discussion focuses on research topics in natural language processing and information retrieval. More precisely, the work covers the tasks related with data selection, filtering and cleaning, morphological normalization and parsing, document representation and similarity computation, and various aspects of data analysis that have all been developed and successfully used in data mining. In the part dedicated to the visualization, the study essentially focuses on the issue of high dimensionality for document representation. Indeed, the high dimensional representations that are produced in the various stages of the text mining process are usually not well suited for a simple and easily exploitable presentation of text mining results which require specific interpretation techniques, tightly connected to the task of document summarization. In addition, the study has identified a clear need for the development of a unified methodology in the field of visualization.

# DELIVERABLE CONTENTS

# LIST OF FIGURES

# 1. Introduction

Text Mining refers to the process of knowledge discovery from large collections of unstructured textual data [Feldman95][Hearst99]. In contrast, data mining was previously studied and practiced mainly in connection with structured data stored in databases. This type of data mining activity, usually known as knowledge discovery in databases (KDD) [Fayyad95], is often defined as a non-trivial extraction of implicit, previously unknown, and potentially useful information from data [Frawley91]. In principle, the process of text mining has a structure similar to the process of knowledge discovery, with, however, some additional requirements and constraints imposed by the specific properties of the processed unstructured textual data expressed in natural language.

In this context, the general goal of this document is to review the main approaches and techniques, especially aiming at the identification of the key research issues in the field of Text Mining.

As natural language is inherently unstructured, there is a need for some pre-processing of the available documents in order to reconstruct the missing data structure. Traditionally, this structure has a form of a feature vector the dimensions of which are associated with the terms extracted from the documents. In the second section we review the document processing techniques that are required for the natural language pre-processing step. In addition to procedures specifically dedicated to the processing of the natural language, such techniques often correspond to some combination of two complementary approaches: information retrieval (for example to select documents to be processed [Strzalkowski99]) and information extraction (for example to extract the features composing the structured representations [Pazienza97][Gaizauskas98]).

Once the targeted structured representation has been produced, the question arises to what extent it is possible to apply standard data mining techniques. We focus on these issues in the document processing section. To this purpose we describe, in the third section, how data mining techniques are applied in the text mining domain.

Another important issue is the high dimensionality of the vectorial representations that are produced and processed at several stages of the text mining process. To deal with this problem, various dimension reduction techniques need to be applied, typically reducing the representation space from thousands into hundreds of features. Dimension reduction is also a key issue for the visualization techniques that are required to support the interpretation of text mining results, as perceptional capabilities of the end users are naturally limited to two or three dimensions.

In the fourth section, we review the most common visualization techniques and discuss their applicability to the textual data.

Finally, a roadmap for follow up research in the domain of Text Mining, along with recommendations for anticipated technological development, is proposed.

# 2. Document Preprocessing

As mentioned in the introduction, Natural Language Processing (NLP) is the main focus for the document preprocessing phase. In this document, we have adopted the generic information extraction model presented by [Hobbs93] dividing the pre-processing of documents into a sequence of distinct natural language processing tasks. The required NLP techniques involve both statistical and machine learning approaches [Charniak93][Daelemans02] [Manning99], often extended by artificial intelligence and information theoretic techniques [Rathnaparkhi98]. they are described in the subsequent subsections.

Finally, as far as evaluation and diffusion of these techniques is concerned, it is important to notice the significant contribution of competitive research evaluation campaigns, such as the Message Understanding Conferences (MUC) [Poibeau03], the Text Retrieval Conferences (TREC), organized annually by NIST and DARPA [TREC], the Document Understanding Conferences (DUC) focusing on Text Summarization [DUC], or the SENSEVAL campaign evaluating Word Sense Disambiguation systems [SENSEVAL]. The important impact of international conferences such as the Conference for Natural Language Learning [CoNLL] and the conferences and workshops organized under the auspices of the Association for Computational Linguistics (ACL, EACL) also has to be stressed [ACL].

## 2.1. Data Selection and Filtering

Data selection and filtering aims at performing a first rough reduction of the vast amount of documents available from numerous information sources in order to avoid the overload related to the rather computationally intensive pre-processing and mining processes.

In the text mining field, data selection consists in the identification and retrieval of potentially relevant documents from available data sources. During this data selection step, the main focus is on the exogenous information associated with the documents, usually represented by explicit descriptive metadata attached to them, such as keywords or descriptors. The main issues related with the task of data selection are therefore tightly connected with metadata interoperability that has been a subject of recent research initiatives, such as the interoperability frameworks developed within the Dublin Core Metadata Initiative (DCMI) and W3C, the Dublin Core and the Resource Description Framework (RDF) [Lassila99].

In contrast to data selection, data filtering focuses on the endogenous information (i.e. the actual content of documents) to evaluate the document relevance. The endogenous information is sometimes denoted as 'implicit metadata'. Basic concepts for textual data filtering are described in [Oard97]. [hull96method] explores some advanced method combination to improve the filtering performance but faces problems due to the small amount of training data.

Compared to the tasks in information retrieval, data filtering is quite close to the problem of document routing, where the focus is rather on the documents in the data source and their changes than on the queries.

Document routing refers precisely to the definition as accepted at the TREC-2 session, which distinguished two text retrieval tasks: (1) the task with an invariant document collection and ad-hoc queries (notion of a traditional search engine), and (2) the task with invariant (routing) query and a document stream. Document ranking is then extended by an additional information, notably the document age and time of integration into the collection, preferring newly arrived and presumably more up-to-date (i.e. timely relevant) documents.

As for performance evaluation the traditional IR measures such as precision, recall and their variants (e.g. the F-measure, the E-measure, non-interpolated and interpolated average precision and the average precision at recall level) are used [Yang99b]. One of the important current research issues is the definition of efficient relevance metrics that are applicable on large volumes of textual data streams.

## 2.2. Data Cleaning

The task of data cleaning is to remove noise from the textual data in order to improve its quality. This goal is also often referred to as "avoiding the GIGO" (Garbage-In-Garbage-Out) effect. Noise can be the consequence of various error sources, leading for example to data inconsistencies and outlying values. The importance of data cleaning also significantly increases when data comes from multiple heterogeneous sources, typically when transformed from one data structure into another or, as it is the case in text mining, when the associated structure is being created out of unstructured data. Among the important data cleaning tasks that are especially relevant in the scope of text mining, one can cite [Rahm00]:

- Spelling error correction

- Reconstruction of missing accentuation

- Letter case normalization

- Abbreviation normalization

- Language identification (to filter out parts that are not in the processed language(s))

- Production of (meta-)linguistic information (PoS tagging, named entity tagging, identification of syntactic roles ,…)

Spelling error correction currently involves several algorithms [Kukich92], usually relying on the notion of edit distance, also called the Levenshtein distance [Schulz02]. A weighted version of the edit distance using weights based on error probability estimations can also be applied. For efficient implementations, spelling error correction using Finite State Automata (FSA)[Oflazer96] can also be considered. An interesting approach applying different metrics for lexical analysis is presented in [Calude99].

Letter case normalization and abbreviation normalization have been reviewed in [Mikheev00]. Various rules can be used, for example capitalized words are expected in two distinguished cases, (1) as indicators of proper names and (2) at special positions in text, such as at the beginning of sentences, quotations, after a colon, etc.

Language identification can be modeled as a text categorization problem, where the categories represent the considered languages. Several methods have been developed for the language identification problem. Simple approaches include the small word technique and the n-gram technique [Grefenstette95]. The Small word technique is based on the idea that common words (determiners, conjunctions and prepositions), which are often small words, are also good indicators of the language used. The small words appearing in a text whose language is to be identified are assigned appropriate probabilities with respect to predefined list of languages, and the probability for a document of be in a given language can then be determined from these probabilities. The n-gram technique (bi-gram, tri-gram, etc.) is based on frequencies of n-grams appearing in a particular language. Various methods that successfully implement these techniques have been investigated, such as the Rank Order, Monte Carlo methods and methods based on Relative Entropy [Poutsma01].

In addition, when processing multiple heterogeneous data sources, data integration techniques are applied in order to obtain a homogeneous data input (including resolution of value conflicts and attribute redundancy[1]). The general task of data integration is defined in more details in the domain of document warehousing [Sullivan01].

## 2.3.   Document Representation

The most common document representation model relies on the notion of feature vectors mapped into n-dimensional vector space (Vector Space Model). In the simplest approach the dimensions of the full-scale feature vectors are associated with the words extracted out of the documents (collection vocabulary). This representation, often referred to as the "bag-of-words" approach, although very easy to produce, is not considered to be optimal for several reasons. One of its main drawbacks is the high dimensionality of the representation space (which grows with the size of the vocabulary used). Notice that the resulting representations with dozens of thousand of dimensions put severe computational constraints on the text mining process. Some dimension reduction is usually performed, leading to the selection of features strongly representative of the content of the documents. The criteria used for feature selection are usually based on word frequency[2] or on more sophisticated selection methods relying on criteria such as chi-square tests, information gain or mutual information [mladenic98feature]. Varying importance for the individual features (integrating for example their discriminative power for the documents in the processed collection) can be taken in account through various weighting schemes, as it is done in the IR field with, for example, the tf.idf (term frequency x inverted document frequency) weighting scheme or some of its variants including Rocchio weighting or Ide weighting [Monz01].

The bag-of-words representations that can be easily generated are used for many content sensitive tasks. However, more sophisticated representations are being investigated for more complex tasks, including more structured semantic models. For example, the use of ontologies has been suggested in [Hotho01], where related terms such as synonyms can be aggregated resulting in a reduced document representation space. The resulting representations are then more oriented on concepts rather than on just words.

When creating the bag-of-words representation  the system relies on a word segmentation algorithm to extract each feature of the vector. Segmentation of text can be more complicated than detecting word delimiters as most language contains compounds expressions. Thus features in the vector should reflect the existence of such expressions and segmentation needs more complex techniques.

---

[1] An attribute is redundant if it can be derived from other attributes.
[2] The distribution of word frequencies in natural language texts tends to follow the Zipf's law with many *hapax legomena* (words occuring only once in the text collection).

In general, segmentation will also rely on lexicon and ontologies but more advanced techniques can be used. Noun-Phrase analysis [evans96nounphrase] extracts meaningful sub-compounds whereas multi-word conflation [klavans97natural] can be used to link morphosyntactically related terms.

In the Distributional Semantics model for Information Retrieval (DSIR) the vector space model has been extended by external semantic knowledge investigating co-occurrence patterns derived from a large corpus [Besancon02a]. The semantics is then defined by a set of contexts in which the particular word appears in a corpus. The incorporation of the semantic knowledge allows to better identify the topicality and to more precisely compute the topical similarity between compared textual entities. An additional refinement of a proposed model by incorporation of the Word Sense Disambiguation technique has been presented in [Besancon01], where the co-occurrences are observed among semantically equal concepts, rather than between linguistic units, using a Markov Random Field (MRF).

Finally, the Generalized Vector Space Model was developed for multi-lingual document representation [Carbonell97][Yang98]. The appropriateness of the vector space model in multi-lingual environments is discussed for example in [Besancon02]. Evaluation of Cross Language Information Retrieval (CLIR) was performed in the scope of TREC-6 [Gaussier98].

## 2.4. Morphological Normalization and Parsing

Morphological normalization refers to a group of natural language tasks that aim at the production of canonical surface forms. More precisely, the basic task is to extract the word root from its inflected or derived morphological form by applying a set of rules to strip identified affixes.

This basic form of morphological normalization is widely known as stemming. Among the renown stemming algorithms, the most frequently cited are the Lovins stemming algorithm (the first such algorithm in existence), incorporating a table of 294 word endings that are processed in a single pass [Lovins68], the Davson stemming algorithm that operates with a more complete list of 1200 word endings [Dawson74], the Porter's stemmer, iterating five times over a set of 1200 English suffixes, using a different table at each iteration [Porter80], the Lancaster (Paice/Husk) stemming algorithm [Paice90], or the Krovetz stemming algorithm, which is a modification of the Porter's stemmer by checking the word against a dictionary before each step [Krovetz93]. In general, these algorithms usually contain some additional conditions in order to ensure that the obtained root remains consistent (to take an illustrative example, preventing the word "king" to be inappropriately shortened into "k"). The stemming approaches have evolved into more complex techniques referred to as lemmatization, incorporating some additional criteria, notably by exploiting the identification of the word's part-of-speech category.

The finite-state automata technology enables to define re-usable regular expressions that generalize over language patterns. Finite-state transducers (FST) [Morhi96fst] were developed to enable the description of regular relations between the morphological base and the corresponding expression in some morphological form. Since these relations are regular, they allow the automated morphological normalization computation [Chanod96]. An example system using finite-state approach is the well known FASTUS system [Appelt93]. Incremental robust parsing (e.g. the XIP parser recently developed at the Xerox Research Center, Europe) and the use of finite state transducers for shallow syntactic parsing was studied and illustrated with examples of applications in [Ait-Mokhtar97].

Parsing aims at assigning some syntactic structure to a morphologically normalized text. It mainly includes text segmentation, sentence chunking into smaller syntactic units such as phrases or syntagms, and the production of syntactic relations between the identified units. Approaches to parsing based on shallow techniques have reached good results for most of the above mentioned tasks, however more sophisticated techniques of parsing as well as incremental robust parsing are still investigated [Ait-Mokhtar97].

Parsing often relies on metadata added to the newly normalized text to construct the syntactic structure. This metadata separates basic syntactic information from the morphological form of text segments. Extraction of such information is often related to as Part of Speech (POS) tagging. POS tags for each item can be extracted from lexicon or ontologies or inferred by statistical methods. Statistical methods like the well known Brill tagger [brill95transformationbased] rely on n-gram or full phrase content [even-zohar00classification] to provide contextual information. [dehaspe97mining] proposes to apply a database mining techniques to produce association rules describing the POS tag sequences.

In general, parsing techniques are divided into two main families of approaches: probabilistic and non-probabilistic. Probabilistic approaches include techniques such as memory-based shallow parsing [Daelemans99] or statistical decision trees [Magerman94].

Parsing is frequently based on grammars to describe the syntactic rules for each language. Parsing using contextual or non contextual grammars is a problem that has been studied in a larger context than natural language processing and efficient parsing algorithms are available. However, simple grammars forms like the Greibach Normal Form are not always sufficient to preserve the original structure of natural languages. Tree-Adjoining grammars [joshi97treeadjoining][nurkkala94parallel] aims at providing more complete grammar rules. Other techniques help to control the ambiguity by adding constraints to the POS tags before the parse [blache-proof].

An often discussed research topic is the use of deep (full) parsing as an alternative to shallow approaches. However, deep parsing has not yet been developed in full extent, partly because of the fact that shallow parsing techniques perform well enough for a large number of applications. Nevertheless, deep parsing is advocated, for example in [Uszkoreit02], emphasizing that there is a real potential for improvement, especially for applications requiring the resolution of more complex linguistic problems such as anaphora resolution. Attempts to integrate shallow and deep parsing have also been recently undertaken and are reported in [Daum03][Cryssmann02] [Frank03].

## 2.5. Semantic Analysis

One of the main objectives of semantic analysis in the domain of text mining is to resolve semantic ambiguities, for example generated by the presence of synonymous and polysemous expressions in the documents. In this perspective, the main areas for semantic analysis are word sense disambiguation (WSD) [Veronis98] and anaphora resolution [Mitkov02].

An interesting use of semantic analysis is the one made by Dassault Aviation [faure98acquisition] and other engineering companies [castell95filtering]. They use NLP techniques to control the complexity and ambiguity of specifications to ensure an errorless reading by the engineers during the implementation phase. Using knowledge bases and semantic analysis, they can ensure that the specified requirements are non-ambiguous and are not redundant.

For the co-reference/anaphora resolution problem, one of the important current open questions is whether such kind of resolution can be performed purely with syntactic tools or if it also requires an integration of semantic or even pragmatic knowledge. A thorough scientific analysis in the domain of co-reference and anaphora resolution can be found in a monograph recently published by [Mitkov02].

In word sense disambiguation, the context of the expression to disambiguate is analyzed in order to assign the appropriate word meaning. The concepts have been introduced as shared word meaning independent from the lexical realization. Concepts are essentially semantically unambiguous and the main issue of WSD can therefore be formulated as the search for a mechanism that reliably assigns concepts to words. [Diab02Unsupervised] proposes an interesting solution to the WSD task using parallel corpora and assuming that words having the same translation share some meaning.

Concepts are organized in ontologies providing means for encoding background knowledge for natural language processing at the level of inter-concept relationships and connections to a shared vocabulary.

In most natural language processing tasks there is a need for structured lexical bases to provide information on language or domain specific terms.

Simple (commercial) lexicon providing POS tags or stemming information are available for most of the languages. In an effort to help WSD and other complex NLP tasks, more complex lexical structures are explored. Ontologies [Sowa00Ontology] provide a mean of storing complex information [fellbaum98towards] on words meaning and context. Public domain ontologies like WordNet and semantic web infrastructures [fensel01oil] permit to perform common tasks.

However getting an accurate and complete linguistic database is always a challenge as its creation is mainly a human task. Hence some efforts are being pursued to automate lexicon creation using machine learning[habert98extending][faure98acquisition] or to construct multilingual lexicon [Calzolari02towards].

Latent semantic analysis (LSA), also often called the Latent Semantic Indexing (LSI), is another frequently used method to reduce the dimensionality of a vector space. The general idea of LSA is to apply a singular value decomposition (SVD) to the matrix the columns of which are the feature vectors corresponding to the representations of a document collection and to decompose it into matrices with lower dimensions. The main advantage of LSA is that it allows coping with semantic issues such as synonymy, polysemy and term dependence [Rosario00]. Most often, the tools applying LSA use the standard tf.idf weighting scheme for the production of the feature vectors. However, it has been observed that with this weighting scheme, the low-frequency terms are underestimated, whereas the high-frequency terms are overestimated; therefore this simple method does not seem to be optimal. There are alternatives to LSA, for example the latent Dirichlet allocation approach that generalizes and improves previous models (Naïve Bayes, pLSI [Hoffmann99] models). An implementation of LSA without SVD is presented in [Hastings99]. Other alternatives to LSA include for example the Linear Least Squares Fit (LLSF), developed by [Yang94], that uses multivariate regression model learning [Deerwester90], Iterative Residual Rescaling (IRR) presented in [Ando01], or novel statistical approaches based on co-occurrence counts, such as the PMI-IR (Pointwise Mutual Information) that integrates an unsupervised algorithm for the processing of synonyms, or the distributional semantics defined in [Rajman99].

The second important objective of semantic analysis is to provide ways of assessing topical proximity between documents, or other textual units (paragraphs, sentences, ...). Issues related with semantic proximity were intensively studied in the field of Information Retrieval (IR). IR techniques assume that the 'true' document relevance with regard to some user query is a boolean value independent from other documents. In the automated approaches, the relevance between a document and a query is measured by their similarity. The existence of various representation models lead to a variety of similarity measure techniques [Rajman98]. Frequently used (dis)similarity measures for the vector space representation are:

- Generalized Euclidian distance

- Chi-square distance

- Cosine similarity

- Ad-hoc coefficients (e.g. Dice coefficient, Jaccard coefficient)

- Measures based on relative entropy.

Another set of similarity measures are compared in [Shang02], namely the Cover Density Ranking (CDR), where relevance computation is based on phrase occurrence; the three-level scoring method (TLS) that mimics manual approaches to evaluation and the Okapi Similarity measure that considers other factors than the term occurrence frequency, such as, for example, the length of the document under evaluation and the average document length in the whole collection [Robertson94]. A similarity measure based on distributional semantics was presented in [Rajman99]. This approach relies on the hypothesis that there is a strong correlation between the distribution of a term in a collection and its meaning.

# 3. Data Mining techniques for Textual Data

In the most general sense, data mining is an analytical process performed on a dataset and using techniques based on statistics, machine learning, information theory and artificial intelligence [Berthold99][Hand01][Hastie01]. Traditionally, data mining techniques comprise classification, clustering and discovery of association rules, with the general purpose of describing, estimating or predicting some set of interesting events. Depending on whether the considered events are present or future, distinction is made between descriptive data mining and predictive data mining [Weiss98]. Text mining essentially belongs to descriptive data mining whereas data mining applications in domains such as business and finance are most often examples of predictive data mining [Berry97][Berson99]. However in connection with text mining, predictive techniques such as trajectory identification and trend analysis are also investigated.

## 3.1. Clustering techniques

Clustering is one of the core data mining techniques. It refers to an unsupervised learning process where individual items are grouped on the basis of their mutual similarity or distance. In the scope of text mining, the clustering techniques are used for the document selection task, as well as for the visualization of results, for example in situations where it is more convenient to display groups of items to make the interpretation of the results easier.

Clustering techniques, which have been extensively studied in fields such as information retrieval, can be decomposed in the partitional and the hierarchical approaches. In partitional clustering, the number of clusters is given as an initial value and remains unchanged over the iteration of the clustering process; in the first step the clusters are chosen randomly and are then iteratively improved; in hierarchical clustering the number of clusters is adjusted during the iterations.

The first clustering algorithm was the "k-means" algorithm [Hartigan75]. Its principle is that each cluster is represented by the mean of all its members and serves as a basis for an iterative cluster re-grouping. Other statistical representations of central tendency can be used instead of the mean, for example the median or the mode. Other algorithms belonging to the same family as k-means have been developed (e.g. PAM - partitioning around medoids). An implementation of these algorithms can be found for example in the CLARA [Kaufman90] or the CLARANS systems [Ng94].

Within the database community, another approach, known as density-based clustering, has been developed [Han01]. The core of the density-based approach is to incorporate two additional parameters – the maximum radius of the neighborhood (Eps) and the minimum number of points in an Eps-neighborhood (MinPts). These two parameters are combined in such a way that too small and too large clusters are discarded. One of the important properties of density-based clustering is its stability and its ability to discover arbitrarily shaped clusters.

In hierarchical clustering, clusters are created by iterative merging (agglomerative clustering) or splitting (divisive clustering) of previously identified clusters. The process of hierarchical clustering therefore leads to the creation of a dendrogram - a tree of clusters - allowing to adjust the clustering granularity according to the particular data mining task. The hierarchical approach is thus considered as more flexible than partitioning clustering. Some examples of hierarchical clustering algorithms are the BIRCH [Zhang96], CURE [Guha98] or ROCK [Guha99] algorithms.

In addition to the above mentioned classical techniques, other alternative approaches have been developed for clustering; for example fuzzy clustering [Bezdeck84] or clustering based on the expectation maximization algorithm [Dempster77][nigam99text].

[Kraft03Textual] uses fuzzy clustering of web documents to determine user profiles. The clusters centers represent the terms most looked at by the user and can be used to establish profiles potentially useful for personalized marketing on the web.

To enhance the performance of the clustering process, some background knowledge can be included. As suggested by [Hotho01], a significant amount of background knowledge that could be profitably used for semantically based text clustering is encoded in ontologies (resulting from a collaborative production process often referred to as the "ontological commitment").

As, depending on the technique used, the number of clusters might need to be specified, one of the important open questions in the field of clustering is to determine the optimal number of clusters for each particular case, and to analyze to what extent this number depends on the nature of data that is being processed. An approach for finding the optimal number of clusters, based on a Monte-Carlo method, has been proposed by [Smyth96].

Other important questions (as pointed out in [Roth02]) are to decide (1) how the appropriate feature vectors are selected; (2) what cluster structures are indeed of interest and (3) what clustering algorithm is appropriate with regard to the anticipated structure? These questions are open research issues and are often referred to in the context of the cluster validation problem.

Cluster validation raises the problem of the quantitative evaluation of the clustering results, possibly with some measure independent from the clustering technique used. Potential cluster quality measures such as cluster stability, cluster compactness, or inter-cluster separation can be quantified [Maulik02] with cluster validation indices such as Dunn's index, Davies-Bouldin index, Calinski-Harabasz (CH) index, C-index, or Xie-Beni index (compactness and separation validity function for fuzzy clustering).

One of the main problems with the iterative clustering algorithms is that they tend to converge towards local extrema (of the criterion used to measure the quality of the clustering), rather than to the desired global one. As the high-dimensional spaces are usually very sparse, the quality of resulting clusters then strongly depends on the initial values used to initiate the iteration and rarely reflect the global characteristics of the data under investigation. Several approaches that attempt to cope with the problem of convergence towards local extrema have been developed. One possibility is to run several iterations starting with different initial values and to select the best output out of all the runs performed; another possibility is to apply some method to choose a convenient initial configuration. Other alternatives are the use of the Monte-Carlo method as in [Williams00] or in [Maulik02]. The latter adopts Simulated Annealing method and iterates over candidate states by either accepting or rejecting the proposed state at each step using the Metropolis criterion known from the random Monte-Carlo simulations. If the iteration steps are small enough, the process is claimed to converge to the global extremum.

A general overview of clustering techniques and related algorithms is presented for example in [Halkidi01] that also identifies several open research issues in the area of clustering, such as uncertainty handling, visualizing of multi-dimensional clusters, incremental clustering and constraint-based clustering.

## 3.2. Classification

Text classification refers to a supervised learning process aiming at assigning pre-defined classes to documents [Sebastiani02]. This classification process can be formalized as the assignment of boolean values to each of the (document, category) pairs, based on some selected criteria.

Classification is often used when a decision has to be taken about a document or a set of document, the destination classes being the different choices to be taken. For example an agent like Personal WebWatcher [Mladenic96Web] will try to classify the hyperlinks in the document between "positive" and "negative" documents regarding the user preference. Such classification is based on the textual features extracted from the different documents.

A large number of approaches has been considered for classification [Sebastiani02]. Among these, the traditional statistical classifiers, the k-nearest neighbor (kNN) [Yang99b] and Naïve Bayes (NB) [Shivakumar00], are considered to have the best performance/complexity ratio. They usually exhibit good performance while remaining fairly simple and efficient to implement. Support Vector Machines (SVM) are another classification method introduced by [Vapnik95] and further examined by [Joachims01]. The goal of SVM is to segment the representation vector space by planes in such a way that members of different classes are separated in the best possible way. One of the important differences of this technique with respect to others is that it only considers a selection of the closest vectors – the so called support vectors. Several implementations of the SVM algorithm are available (for example SVM light [Joachims98]). An algorithm for training SVM called the Sequential Minimal Optimization (SMO) algorithm is described in [Platt98]. [he00comparative] evaluates some classification method for Chinese; raising the problem of correctly segmenting Chinese text but demonstrating that actual techniques can efficiently be applied to languages more complex than English.

Among the vast amount of other possible approaches to text classification, one can cite decision trees (ID3, C4.5, CART), decision rules, neural networks, bayesian networks, genetic algorithms, distribution estimation [nigam99using] and example based classifiers. A comparison of some of these approaches was undertaken by [Yang99a]. Widroff-Hoff and Exponential Gradient (EG) algorithms described in more detail in [Lewis96], are particularly suitable for larger feature vectors using the tf.idf representation.

## 3.3. Relation Extraction

The usual bag-of-words representation does not capture any relations between individual features in the vectors, and this fact may lead to a substantial loss of important information. In order to discover these relations, various techniques have been suggested since the MUC-7 evaluation conference, where this extraction task was first defined (more precisely, the task was the extraction of relations between pairs of entities such as person-employer, market-product, organization-location). Approaches usually considered are typically based on morphological analysis and shallow parsing. Relation extraction also includes event extraction, which specifically focuses on extraction of entities related to some event, such as, for example, a conference, workshop, etc.

Relations between entities that do not occur in the same sentence (it is estimated that such relation represent about 10-20% of all relations), can also be considered, as it is the case, for example, in the cross-sentence feature model [Miller98].

Integration of co-occurrence patterns into the bag-of-words model was investigated by [Jing99]. Semantic relations discovered through feature co-occurrence are tightly connected with the issue of association rules discovery, where the associations between sets of features are being discovered. The general process of association rule discovery usually consists of two main phases: (1) the generation of candidate feature sets (often called the frequent sets) that co-occur in the available data with a frequency exceeding a predefined threshold, and (2) the extraction from the set of all the possible associations between the entities present in a given frequent set of all the ones that satisfy some predefined quality measure, such as, for example the confidence or the lift [Agrawal96][Feldman95][Feldman96][Rajman97].

When the association rules have been discovered, general knowledge and database mining techniques like the reduction of the number of rules [toivonen95pruning] can be applied.

[loper00applying] uses semantic relation extraction to provide stronger IR. SQUIRE extracts semantic relations in the query sentences and compares them to an existing database of indexed documents. This retrieval technique offers better results than the common ones that use only query term occurrences to rank the documents.

## 3.4. Entity Extraction

The entity extraction task aims at assigning some pre-defined labels to the textual entities (words, compound words, expressions, …) that hold a given interesting semantic property. Typical examples of entities are company names, person names, dates, phone numbers, or prices. These entities are often partially identified by lists provided by an external knowledge source. They are also often structurally characterized by set of descriptive patterns. In [Hearst92] these patterns were coded by hand, which is a very time-consuming task. An automation of the pattern identification has been recently addressed in [Bergstrom00] by applying machine learning techniques, namely the genetic algorithms. A frequently cited system is the Alembic entity tagging system [Day97].

[Takeuchi02use] make the use of SVM to extract entities in the molecular biology domain. They try to extend the usual techniques to render extraction of instances of conceptual classes easier. This technique is used on abstracts in the domain of biology to extract entities like protein name, genes, etc…

The entity extraction task was originally formulated as named entity recognition in the scope of the MUC-6 evaluation conference. Named entities were the following: person, organization, location, date, time, money and percent.

Typically, the entity extraction is preceded by a group of NLP operations, namely list lookup (selection of name candidates), tagging of proper names, name parsing and finally name matching.

The named entity extraction has been used for example in TopCat [Clifton01] (Topic Categories), a recently developed NLP-based technique for topic identification. One of the questions raised is, whether the lists of names could be replaced by some annotated techniques while maintaining the performance at the same comparable level.

Recently, entity extraction was defined as a shared task at the CoNLL'02 evaluation conference. Twelve systems have been tested on two languages (Spanish and Dutch), achieving the F-values of 81.39% for Spanish and 77.05% for Dutch. These values were reached by system described in [Carreras02], which uses external knowledge sources allowing a reported performance improvement of 2%.

Finally, term extraction [Feldman98] is a technique that allows discovering in the texts compound expressions that have a significant meaning. As the meaning of the "true" compound expression is usually not easily derivable from the individual meanings of the atomic words, the discovery of this type of compounds brings clean added value to the text mining process.

## 3.5. Chronological data analysis

Data analysis and knowledge discovery in database are generally based on static data. For example, static data analysis often consists in finding association rules between properties of the data and exploiting them for prediction tasks. However some works have demonstrated the need for analysis of time-related data [Saraee][Saraee95]. An example of time-oriented analysis is a search for extended association rules where the temporal nature of the antecedents and consequents in the rules relations is taken into account. Such rules are generally known as "Sequence rules" or "Sequential patterns".

Chronological knowledge mining in textual corpus has not yet been much explored. However some interesting concepts and applications have emerged.

Actually, the main task in this field is described as "Topic Detection and Tracking" (TDT) which have been defined by a DARPA pilot study [Allan98] in the following way: TDT consists in detecting the emergence, in a stream of textual data, of new topics and in following their evolution. TDT is decomposed in three tasks:

- segmenting a stream of data, especially recognized speech, into distinct stories;

- identifying those news stories that are the first to discuss a new event occurring in the news;

- given a small number of sample news stories about an event, finding all following stories in the stream.

The segmentation task consists in extracting topically homogeneous blocks from a serial data stream. Different approaches are described in the above cited paper. For example, Hidden Markov Model (HMM) can be used to predict transition between topics; HMM are equivalent to finite state machines with a probability of transition between each state depending on various features such as the previous state or the input observations. For segmentation, the hidden states are the topics and the observations the selected text tokens (e.g. words, sentences). The transition probabilities are learned on a training corpus containing annotated data. The main idea in all the presented methods is to detect the changes in the occurrences of content bearing words.

Once the corpus segmented, the next task is to tag each segment as a new or old story. Such a task can be seen as a classification of the segments resulting of the previous task into two distinct sets. However, the method presented in [Allan98] is based on clustering: term frequencies are computed for each story, which are then clustered on the basis of their lexical proximity. Hence each cluster will represent a novel event and the oldest story in each cluster will be a "new" story.

Some alternative methods have also been considered. For example [Rajaraman01] and [Pottenger01] use neural networks to extract patterns in the stories and classify them; this is a black-box approach, as the story are clustered by the network according to its architecture and the used patterns can only be extracted by separately studying the documents themselves. [Uchimoto97] provide a simpler method to detect topic changes in newsgroup's threads: each article is represented by "clue-words" and the topic change is detected by looking for transitions in words frequencies.

Once a topic is detected, tracking it in the subsequent stories is a task similar to document routing in Information Retrieval: given a set of query, a document should be routed to the most relevant query. However, TDT does not provide queries but a set of initial stories, each of them describing a different event; hence the subsequent stories should be associated to the most relevant event described by the initial stories. This can also be seen as clustering, where each initial story represents a cluster and all the following story are clustered to the nearest cluster.

Once a TDT task has been performed, the textual data is structured along a temporal axis and, using other IR methods, some trends can be extracted and analyzed using specific query [lent97discovering] and visualization techniques. [Baldwin98] propose fuzzy logic using rules based on natural linguistic terms describing trends. He can then produce a "glass box model" of the series using the trend described in the stories.

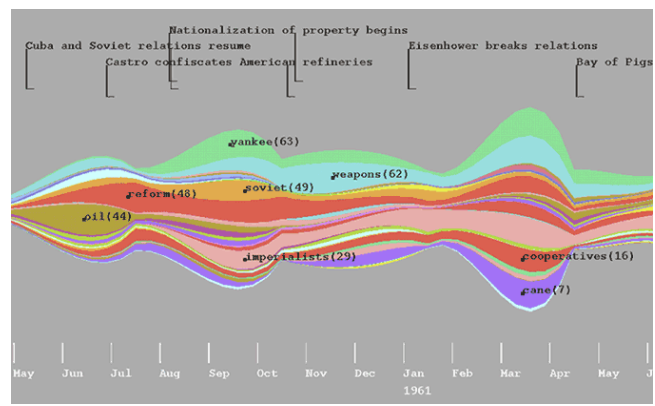

**Figure 1. ThemeRivertm visualization of topic variation over time (© Pacific Northwest National Laboratory).**

An interesting application of TDT is the study and prediction of the effects of events extracted form textual data on time series. For example, in financial analysis, there is a great interest in knowing the influence of news stories on the financial market (and reciprocally). [aenalyst] identifies trends in time series using numerical methods and then align these trends with news article and use a language model to detect emerging events in the textual corpus useful to make predictions. Such a system can also be used to recommend interesting financial news to analysts. Time series analysis based on text solely (without additional numerical data) might also be considered.

# 4. Visualization of results

In the domain of text mining, visualization essentially faces two main problems, related to the nature of the (textual) data that needs to be visualized:

- the representations most often use vector spaces of very high dimensionality; in addition to the dimension reduction techniques that are applied during the pre-processing phase, specific visualization techniques need to be used to make the high dimensional representation visible in low dimension (typically 1,2 or 3D) spaces; these techniques will be reviewed in section 4.1.

- when clusters of documents are visualized, additional information needs to be produced and associated with the cluster so that the user can have some idea of its content; summarization techniques can be used for this purpose and will be reviewed in section 4.2.

## 4.1. Visualization Techniques for multi-dimensional data

A comprehensive classification of visualization techniques has been provided by [Keim02] who presents an overview of available data visualization techniques, grouped in the following main categories:

- geometric projection (scatterplot matrices, prosection views, parallel coordinates, hyperslice, PCA, FA, MDS,...)

- pixel-oriented (recursive pattern technique, circle segments technique, spiral- & axes-techniques)

- icon-oriented (Chernoff's faces, stick figures, shape coding, color icons, tile bars)

- hierarchical (cone-trees, treemap, Wenn-diagram, dimensional stacking)

- graph- and distortion-based techniques (fish-eye, hyperbolic trees, table lens).

An alternative taxonomy have been presented in [Chi00] that classifies the techniques by their data type and "processing operating steps".

Geometric projection focuses on a geometric transformation of data aiming at projecting it into a lower dimensional space. Dimension reduction using projections (also called projection pursuit [Keim02][Huber85]) aims at geometric projection of a high dimensional data onto a space of low dimension (typically 2D or 3D). Multivariate data analysis techniques [Lebart98], such as the principal component analysis (PCA), the factor analysis (FA), multidimensional scaling (MDS) [Sammon69][Anderson58], and projected clustering [Aggarval99] can be applied.

Principal component analysis (also called Karhunen-Loève transform or Hotelling transform) seeks for projections based on linear combinations of a selection of the original variables (dimensions) that maximizes variance.

Multidimensional scaling (MDS), also referred to as Sammon's Projection [Sammon69], is a multivariate statistical technique similar to factor analysis. It was originally used to visualize results of psychological data reducing the space into two or three dimensions. MDS differs from methods such as PCA by the fact that the distances between points in the original vector space are taken as an input rather than the point coordinates themselves. However, this method was not yet fully evaluated in the domain of text mining.

When the representation vector space is high-dimensional, some of the methods proposed above are time-consuming and sometime infeasible. Ritter and Kohonen proposed an alternative for reducing the dimensionality, where each dimension of the original space is replaced by a random direction in a smaller-dimensional space [Kaski98].

Self-organizing Maps (SOM) are a numerical non-parametric data visualization technique that uses specific neural network architecture (the Kohonen Maps) to perform a recursive regression leading to data dimension reduction. A thorough description of the SOM principles can be found in [Kohonen95][Kaski97]. One of the interesting properties of the SOMs is that they enable the extraction of invariant features from the data set. Some tools based on the SOM approach have been developed, for example WEBSOM, a tool dedicated to the visualization of documents published via the WWW.

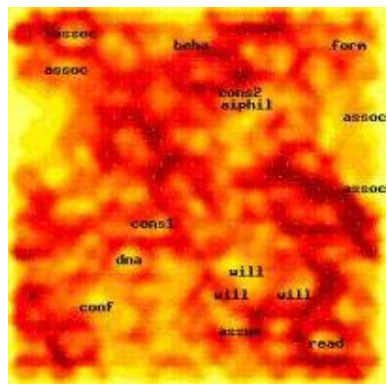

**Figure 2. WEBSOM document map.**

Pixel-oriented visualization techniques map each data item to a colored pixel that is appropriately placed in a space of reduced dimensionality where each dimension is represented in a separate window (recursive pattern technique) or a segment of a circle (circle segment technique). 1D sequences of pixels are mapped onto 2D spaces using for example the Peano-Hilbert or Morton (Z-curve) space-filling curves [Keim00].

Icon-oriented techniques use various visual properties that can be associated with the items to display additional dimensions. Each additional dimension is represented by a particular property of the item, such as shape, size, color and location in the 3D space. The most favorite technique cited is the Chernoff faces technique, where each dimension is displayed as a fragment of a human face sketch. Another frequent technique uses sticks, where values in the additional dimensions are encoded into the stick length and angles between the sticks. However, such approaches allow to visualize about a dozen of dimensions and as such, they often do not meet the high-dimensionality requirements imposed by the vector representations of textual data.

Notice that, in many of these approaches, a common known problem is that the additional dimensions located in separate display items (e.g. circle segment technique or parallel coordinates in 1D ordering) have to be ordered according to some key. One idea presented by [Keim00] is to compute the similarity between the displayed dimensions and place similar dimensions next to each other, employing Euclidean or more complex similarity measures (e.g. Fourier-based).

Hierarchical visualization has been implemented for example in the XmdvTool software package developed at Worcester Polytechnic Institute (WPI) [Rundensteiner02]. It applies another dimension reduction technique referred to as the visual hierarchical dimension reduction framework (VHDR). One of the main advantages of this approach is that it allows the end user to interactively select the appropriate granularity level of display.

Graph visualization [Herman00] is generally intended to display structured data which offers explicit relationship between the corpus items. Still, the graph visualization is used in text mining in conjunction with other IR techniques like the Pathfinder Networks [Chen99vizualization] to offer a structured view to the user and help discovering implicit relations in the corpus.

## 4.2. Visualization in text mining

[Hetzler98] explains the need for visualization techniques in the domain of text mining by the fact that tools are required to help the information analysts in achieving their goals. In particular, the author believes that in text visualization there is a need for a "multi-faceted approach" to improve the analysis. He lists what should be provided to achieve such an approach:

- getting a sense of the major themes in the information and how strongly they are present,

- understanding how themes relate to one another,

- seeing how the information relates to another information collection or to a standard ontology,

- getting a summary of the attributes of the information, such as source, date, document type, etc.

Clustering has already been mentioned as a technique to reduce the document space dimensions. For example, in *Bead* [Chalmers93] similarity between articles is represented by their relative geometric proximity, which leads to a spatial clustering of the documents collection helping the user to discover document's relationships and the potential themes of the corpus. Such a "landscape metaphor" offers an interface that can be easily navigated by the user who can then concentrate on the exploration of the data.

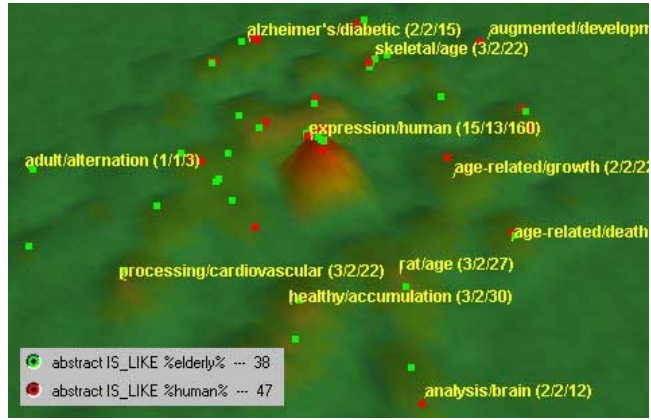

**Figure 3. VxInsight landscape display.**

For Information Retrieval (IR), the main research directions in visualization techniques focus on the display of retrieval results. White and McCain [White97] provide some evaluation criteria for such interfaces and state, for example, that they should provide a clearly perceptible improvement over a simple list display.

This kind of basic display is often extended by the indication of ratings based on various ranking scheme relying on diverse relevance measures (e.g. distance of the document vector from the query vector); however, [Noreault81] demonstrates that these ranking schemes do not offer high improvement over a random sort. Hence some different approaches for displaying the document set have been advocated.

The "Cluster Hypothesis" [Hearst96] states that "relevant documents tend to be more similar to each other than to non-relevant documents" and therefore visual clustering of retrieved documents is an interesting alternative to lists [Au00]; Notice however that this hypothesis is only valid if retrieval is done from a large corpus of documents of sufficient size (it cannot be efficiently applied to small documents such as titles or abstracts).

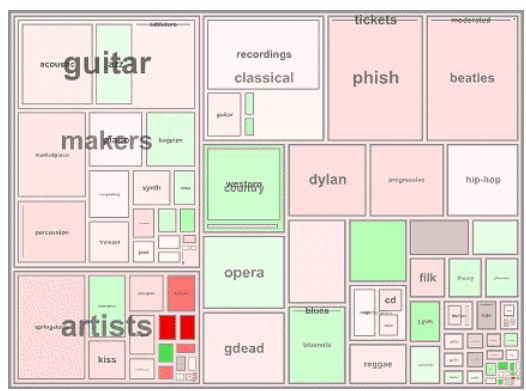

**Figure 4. TreeMap hierarchical clustering.**

As the complexity of the visualization techniques grows (with complex data visualization such as visual clustering or graph display) an increasing problem is the gap with user's mental model. Evaluation of different IR interfaces [Sebrechts99] shows that, for basic users, the text/list view is still the fastest. Some effort [Hearst95][Ogden99] have been done to extend the list interface with visual hint (representing for example document length, query term frequency and distribution) to offer to the user additional information on the intrinsic meaning of the ranking scheme.

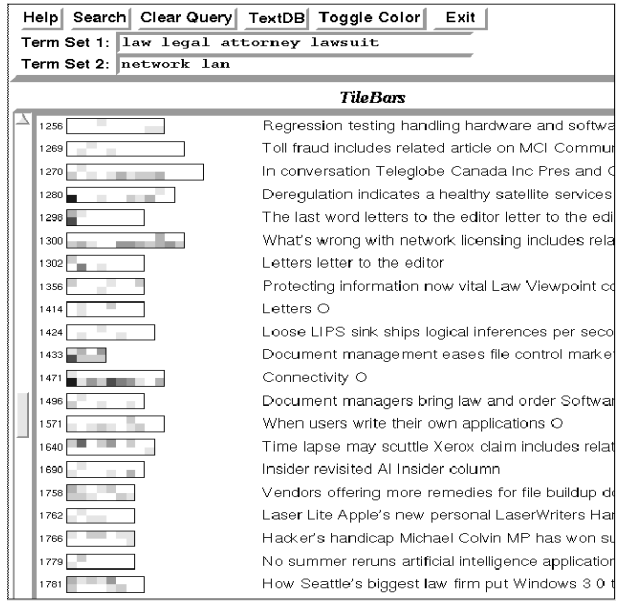

**Figure 5. The TileBar display paradigm. Rectangles correspond to documents, squares correspond to text segments, the darkness of a square indicates the frequency of terms in the segment from the corresponding Term Set. Titles and the initial words of a document appear next to its TileBar.**
**(From [Hearst95])**

Another approach is to offer higher query possibilities to the user. For example [InfoCrystal] provides the user with an extended Venn diagram to visualize the different keyword based boolean queries; other approaches use interactive clustering to create the document cluster dynamically at each refinement of the query [Hearst96][Au00].

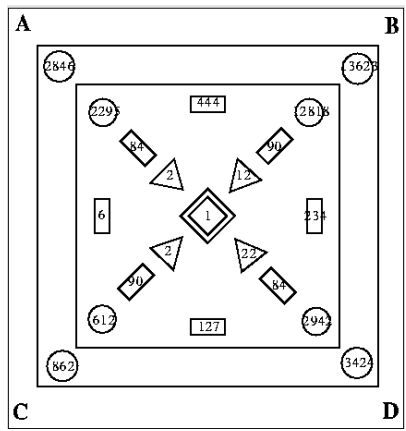

**Figure 6. Sketch of The InfoCrystal (From [Hearst95]).**

The visualization of relations within the corpus [wong99visualizing] is often considered as a good starting point for information analysis. It usually offers a useful insight for the discovery of relations between topics in the corpus. Different types of relations have been visualized: [Chen02a] explores some techniques to display scientific knowledge and relations between researchers using the analysis of citation and graph display techniques, [wong99visualizing] and [Hetzler98a] explore more general techniques to display the relations between documents or topics (hyperlinks, citations, etc...) using visual clustering or matrix display.

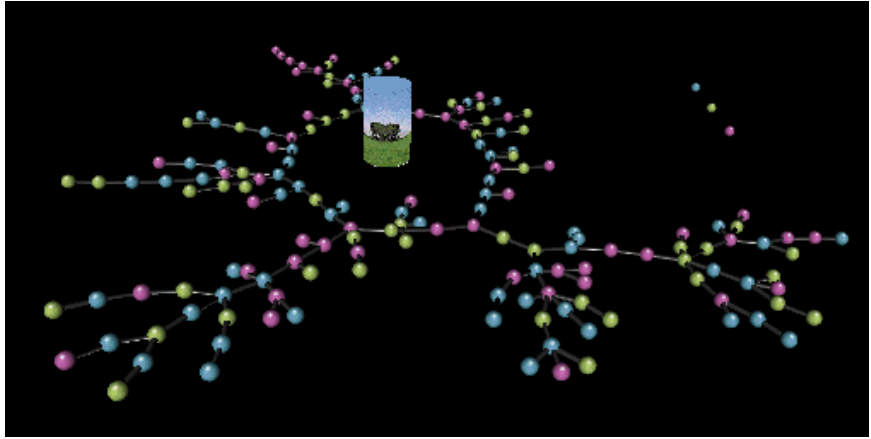

**Figure 7. A "semantic constellation" of an information space. The balls represent documents and their spatial arrangement shows relationships between them.**

Analysis of textual data also includes the extraction of trends or theme evolution over time; however, this field doesn't offer yet a lot of visualization techniques. Still, some interesting approaches have been proposed by [Wong00] or [Havre] to combine statistical analysis and user knowledge to extract sequential patterns.

Some systems provide tools for information visualization but are rather general tools not specifically targeted on textual data mining. For example, tools like [NIRVE] or [VisIt] provide some visual interface to Internet or database search engines and [OpenDX] or [Lyberworld] provide an extended set of tools for generic data exploration.

## 4.3. Text Summarization

When the visualized data represents textual entities or a cluster of textual entities, the visualizations have to be accompanied by some description providing the end user with some additional information about the associated content. One of the possible techniques to produce this additional information is text summarization [Hahn00][Mani01]. In very general terms, text summarization can be defined as a reductive transformation of a source text into a summary text through a content reduction based on the selection and generalization of what is important in the source [Spark-Jones99].

Several well performing systems have thus been constructed in the scope of the TREC evaluation conference, such as QA-tracks, NTCIR, QAC or NIST DUC.

The currently most explored approach for summarization is the extractive approach that essentially builds upon techniques borrowed from the information retrieval field. In this approach, the summary is composed of sentences selected as semantically representative for the document content. An example of an extractive multi-document summarization approach is the TNO system [Kraaij02]. Abstractive document summarization technique is also investigated, and an empirical study comparing the two approaches has been undertaken by [Carlson01].

Among the important research topics that are currently investigated, one can cite the design of non-extractivedocument summarization techniques, language modeling for text summarization, and summarization of non-textual elements (such as diagrams) present in the documents.

Evaluation of summarization systems is still considered as quite open research issue [Mani01a][Goldstein99]. The first coherent evaluation scheme for automated summarization systems was presented at the SUMMAC conference in 1998. Other summarization evaluation actions are the Text Summarization Challenge (TSC) conducted in 2000-2001 in Japan, or the MEADEval project. Various evaluation critreria were defined such as the consistency between documents and summaries for categorization (categorization performed on summaries should produce the same results as categorization performed on the full documents); or informativeness (to what extent does the summary contain answers found in the full document). Traditional IR evaluation criteria (precision, recall, F-score) are often used, but specific evaluation criteria focusing on the summary coherence, informativeness and compression rate are also defined.

Other recently addressed text summarization research topics have been multi-document summarization, multi-lingual summarization and hybrid multi-source summarization. [Chen02][Mani01]. A knowledge-based text summarization has been also addressed by [Hahn99], emphasizing the potential of concepts and conceptual relations as a vehicle for terminological knowledge representation.

# 5. Roadmap for the future research

The specific characteristics of the document processing and visualization as a part of the text mining process were identified and described. There is a large amount of techniques available and many of them are already supported by existing tools. We have provided a general overview of these techniques and identified needs for further developments to cope with the complexity related to the intrinsic properties of natural language. In this section we are going to review the identified open research topics. In general, all identified steps of the text mining process involve a certain information loss as a side effect, which imposes a limit for the quality and reliability of eventually inferred knowledge.

In information retrieval, the currently used methods based on metadata usage do not scale well for the reason that explicit metadata is either missing or not well structured. There are currently two observable trends in the information retrieval research that try to cope with this issue:

■ denoting metadata explicitly in a unified structured way, represented in the first place by the RDF and related W3C standardization efforts,

■ or to target the implicit metadata inherently encapsulated in each item that is a subject of retrieval.

A review of recent research in the field of data cleaning presented in [Rahm00] concludes that there has been so far a little research present in this area, particularly when unstructured textual data is involved, since traditionally the data cleaning techniques focused on processing structured and semi-structured data integration in the scope of data warehousing. Letter case normalization in proper names, as an instance of a data cleaning task, has not yet received an appropriate attention in the scientific community, however, in many cases it can play a significant role for the NLP processing, even to a semantic level. Consider for example the words "Internet" and "internet" referring to a different concept, "Internet" being a sub-classification of "internet". Other examples of this problem have been illustrated in [Church95].

Another task belonging to the data cleaning group is the task of a language identification that is being considered to perform at the limit of theoretically achievable results, where at the average sentence length (11-15 words per sentence), successful cases of language identification range from 93.6% to 100% for the trigram technique and from 97.2% to 99.8% for the small words technique [Grefenstette95][3]. [schultz96lvcsrbased] tries to combine language identification to speech recognition techniques and demonstrates that better results are obtained by using as much linguistic knowledge as possible, but this implies a great amount of processing. The activities in this area now focus on developing methods allowing optimized processing with respect to the processing speed. It has been reported that there is a space for processing speed improvement with no language identification accuracy loss [Poutsma01].

Concerning the document representation, the research focuses very strongly on the Vector Space Model (VSM) and its extensions. [Daelemans02] presented some preliminary result on feature selection using genetic algorithms. They found that the GA-based feature selection does not perform significantly better than the simple feature selection methods [Kool00], however, they pointed out that these findings are not based on statistically significant

---

[3] This test included Danish, Dutch, English, French, German, Italian, Norwegian, Portuguese and Spanish

amount of experiments and state that further research is needed. [Besancon01] , having observed promising first results when applied on IR tasks, states that the MRF model seems to provide a good framework for document representation. He however points out that more thorough research should be conducted in this direction. It is also pointed out that the co-occurrence model could be experimented with in a WSD task itself, which is yet another open area to be investigated.

[Jing99] prompts for more investigation of the IR similarity computation between documents and linguistic entities in the scope of a vector space model. In this area, the latent semantic analysis model is particularly interesting to investigate in more details. The question of optimal dimension to which the eigenvector matrix is reduced into at the singular value decomposition applied within the LSA indexing that was imposed already by [Deerwester90], has not yet been scientifically answered. [Baeza-Yates99] imposes a question, whether LSA remains superior in practical situations with general collections, which remains to be verified". [Rosario97] stated in his LSA overview paper that the LSA could be improved especially for processing of a large collections, which is a key issue for the text mining, particularly the efficient computing of a truncated SVD of extremely large and sparse matrices.

Some evaluation [voorhees99natural] tends to demonstrate that pure linguistic techniques are not always the solution (i.e. in classification or retrieval). The early stage of research in that domain can decrease the performance compared to more generic statistical techniques. Hence, the development of errorless linguistic resources and adapted algorithm is required before the adoption of straight linguistic methods.

Another interesting research topic is the cross-language processing. [Gaussier98] has concluded that cross-language retrieval, as defined at the TREC-6 session, is feasible using an existing technology, but pointed out the following open issues as resulted from an evaluation by NIST: (1) dictionary coverage, (2) translation issues such as translation of unknown words, translations of non-compositional phrases, proper weighting of retained translation alternatives, and (3) stemming issues, notably a proper level of derivational stemming. [Carbonell97] states that there is a need for a future research of the GVSM, particularly in connection to the trans-lingual information retrieval.

Whereas the Part-of-Speech Tagging has been reported to perform at the highly successful level of 95%, other morphological normalization in the text pre-processing task remain an open issue with regard to their performance. Deep parsing has been presented as a potential alternative to shallow parsing by [Uszkoreit02]. In particular, an extension of robust shallow parsing techniques by employing a deep selective analysis is being considered. Such an extension is considered as a non-trivial task, as the two approaches are difficult to synchronize. The principal question here therefore is, whether the deep parsing techniques can achieve the required scalability, or, how the shallow parsing techniques could be efficiently extended to perform well even for complex text pre-processing tasks or, eventually, how these two complementary techniques could be efficiently combined. This includes predominantly the question of their parallel processing synchronization. [Cryssmann02]. The main issue in deep natural language processing is the insufficient robustness and scalability when processing large quantities of unrestricted text.

[Clifton01] identified a problem of topic persistency over time, having in particular the two following issues:

- performance,

- new knowledge;

the latter being considered as more challenging. Clifton also states that another issue is using additional types of information, giving an example of the Alembic project that is working on extracting events. How to best use this information is an open question. For example, grouping events into types may or may not be appropriate. Further work on expanded models for data mining would have significant benefit for data mining of text.

[Maulik02] identifies a need for an extensive theoretical analysis comparing the cluster validation indices, in addition to experimental results. Particularly the evaluation of the validation indexes in combination with various clustering techniques and distance metrics (including other than Euclidean metrics) has to be evaluated.

[Halkidi01] concludes that though cluster analysis is subject of thorough research for many years and in a variety of disciplines, there are still several open research issues. They summarize some of the most interesting trends in clustering as follows:

- *Discovering and finding representatives of arbitrarily shaped clusters. One of the requirements in clustering is the handling of arbitrarily shaped clusters and there are some efforts in this context. However, there is no well-established method to describe the structure of arbitrary shaped clusters as defined by an algorithm. Considering that clustering is a major tool for data reduction, it is important to find the appropriate representatives of the clusters describing their shape. Thus, we may effectively describe the underlying data based on clustering results while we achieve a significant compression of the huge amount of stored data (data reduction).*

- *Non-point clustering. The vast majority of algorithms have only considered point objects, though in many cases we have to handle sets of extended objects such as (hyper)-rectangles. Thus, a method that handles efficiently sets of non-point objects and discovers the inherent clusters presented in them is a subject of further research with applications in diverse domains (such as spatial databases, medicine, biology).*

- *Handling uncertainty in the clustering process and visualization of results. The majority of clustering techniques assumes that the limits of clusters are crisp. Thus each data point may be classified into at most one cluster. Moreover all points classified into a cluster, belong to it with the same degree of belief (i.e., all values are treated equally in the clustering process). The result is that, in some cases "interesting" data points fall out of the cluster limits so they are not classified at all. This is unlikely to everyday life experience where a value may be classified into more than one categories. Thus a further work direction is taking in account the uncertainty inherent in the data. Another interesting direction is the study of techniques that efficiently visualize multidimensional clusters taking also in account uncertainty features.*

- *Incremental clustering. The clusters in a data set may change as insertions/updates and deletions occur through out its life cycle. Then it is clear that there is a need of* *evaluating the clustering scheme defined for a data set so as to update it in a timely manner. However, it is important to exploit the information hidden in the earlier clustering schemes so as to update them in an incremental way.*

- *Constraint-based clustering. Depending on the application domain we may consider different clustering aspects as more significant. It may be important to stress or ignore some aspects of data according to the requirements of the considered application. In recent years, there is a trend so that cluster analysis is based on less parameters but on more constraints. These constrains may exist in data space or in users' queries. Then a clustering process has to be defined so as to take in account these constrains and define the inherent clusters fitting a dataset.*

Usage of large linguistic resources was suggested in [Veronis98], stating that an application of enhanced statistical methods allowed to reach near the limit of what can be achieved in the current framework, prompting for a need of a paradigm change in the research directions in order to allow more significant improvements. This is mainly due to the fact that word senses have yet been hardly well defined [kilgarriff97dont], which implicates a limit for the word sense determination and the WSD task as a whole. The WSD problem could benefit from theories of meaning and from the area of lexical semantics, contextual information and background knowledge, such as encoded in ontologies.

Evaluation methods for document summarization require more research, as stated in [Mani01a] that suggests that summarization corpora may play a valuable role. New areas in document summarization have been identified, namely multi-document summarization and multi-lingual summarization in particular [Baldwin00]. New applications for summarization will define further research directions for summarization evaluation.

Development of evaluation techniques for all the NLP domains is still a need. [King98issues] proposes the adoption of a standard framework to render evaluation development easier.

[Keim02] prompts for an integration of existing visualization techniques and other techniques of statistics, machine learning, operations research and simulation. [Keim00] points out that the visualization techniques per se seem to be ad-hoc solutions without any formal basis, giving an example of pixel-oriented visualization techniques that have a number of unresolved optimization issues.

The taxonomy of visualization techniques has not been agreed upon uniformly. The future work is being considered towards more unified taxonomy and classification of visualization techniques. At the same time, the existing variety of domains, which text mining is one of, leads to a need of meta-analysis of similarities and differences between different data domains [Chi00].

Most of the visualization techniques actually used are not specific to textual data mining. Therefore an effort should be made to adapt the evaluations and recommendations about scientific visualization [mann02scientific] to the more particular domain of textual data mining.

[Rajaraman01] suggested that it would be interesting to integrate domain-specific knowledge into the topic detection and tracking method in order to improve the results of a trend analysis.

# 6. Glossary

| | |
|---|---|
| **Anaphora resolution** | Determination of references to earlier or later words (items) in the discourse that identify the same object. Depending on the Part-of-Speech of the respective words, it is referred to pronominal anaphora, lexical noun phrase anaphora, noun anaphora, verb anaphora or zero anaphora. |
| **Chunking** | Grouping words into phrases and clauses based on their determined morpho-syntactic class, used as a preliminary step to parsing. |
| **Cluster stability** | A criterion for cluster validation, based on variability between clusters created from various initial configuration. A clustering algorithm complying with this criterion is self-consistent. |
| **Cluster validation** | An assessment of data partitioning consistency at the task of data clustering. Several criteria were suggested for cluster validation, namely the cluster stability criterion, cluster compactness or inter-cluster separation. |
| **Deep (full) parsing** | Set of methods assigning a full syntactic structure to a sentence. |
| **Document routing** | Classification of previously unseen documents (typically in a stream) according to user's persistent information need expressed as a query. |
| **Edit distance (Levenstein distance)** | A minimum number of "edit" operations (comprise insertions, deletions and replacements) needed to transform one string into another. |
| **E-measure (IR measure)** | Relative measure derived from recall and precision where various weights of precision and recall can be specified: $E = (1-b2) / (b2/recall - 1/precision)$, where $b>1$ emphasizes precision over recall and $0<b<1$ vice versa. |
| **Expectation Maximization (EM)** | Iterative method for learning maximum likelihood parameters of a model where some of the random variables are observed, and some others are hidden. The expectation maximization algorithm is used to approximate a probability |
| | function, typically by computing maximum likelihood that estimate given incomplete samples. |
| **Factor analysis** | A multivariate statistical technique used to reduce a large number of variables to a few constructs that can be more easily interpreted. To achieve this purpose, this model introduces latent variables, commonly referred to as factors, and posits that the observed variables are determined by these factors. |
| **Feature** | A characteristic word extracted from a document that is used for its representation. |
| **Finite state transducer** | Finite-state device in which the arcs between nodes are composed of pairs, triples, etc., rather than simple symbols. In the scope of Text Mining, FST can be used as a possible technology for syntactic parsing. |
| **F-score (IR measure)** | Relative measure derived from recall and precision as their harmonic mean allowing the comparison of various IR systems. F= (2*precision*recall) / (precision + recall). |
| **Fuzzy** | Assuming uncertainty, employing probability values <0,1> instead of boolean values {0,1}. |
| **Fuzzy clustering** | Clustering allowing partitioning data with no sharp boundary between obtained clusters. A typical example of fuzzy clustering algorithm is the "fuzzy c-mean" algorithm. |
| **Garbage-In-Garbage-Out (GIGO)** | An effect consisting in the fact that that noisy input data extends the probability of inference of inaccurate results. In an extreme case of an invalid data input the resulting output will also be invalid. |
| **Generalized Vector Space Model** | A generalization of the classic vector model based on a less restrictive interpretation of term-to-term independence. |
| **Hidden Markov Model** | A special case of a non-deterministic probabilistic finite state automaton, where inferences are done indirectly as the underlying stochastic process (Markov chain) is only estimated (hidden). |
| **Incremental robust parsing** | Incremental parser consists of a sequence of transducers compiled from regular expressions, where at each step transducer adds syntactic information. |
| **Latent Semantic Indexing** | An algebraic model of document retrieval based on a singular value decomposition of the vector space of index terms. |
| **Lemmatization** | A sophisticated method of a determination of a grammatical root of a word. In comparison with stemming it incorporates additional techniques and knowledge in order to improve the results. |
| **Metadata** | Attributes of data or a document, usually descriptive as author or content, often broken up into categories or facets, typically maintained in a catalog, sometimes recorded according to a scheme like the Dublin Core or MARC. |

| **Monte-Carlo method** | Analytical technique in which a large number of simulations are run using random quantities for uncertain variables and looking at the distribution of results to infer which values are most likely The Monte Carlo method encompasses any technique of statistical sampling employed to approximate solutions to quantitative problems. |
|---|---|
| **Morphology** | A study of the internal structure and the variability of words in language. |
| **Multidimensional scaling** | A method to map objects into points in an Euclidean space, trying to preserve distances between these objects. The dissimilarity of objects is then expressed as a distance between these points. |
| **Multivariate statistics** | A branch of statistics which measures changes in a number of items simultaneously. Some of the most common multivariate statistical techniques are cluster analysis, conjoint analysis, factor analysis and multidimensional scaling. |
| **n-gram** | Any sub-string of a length n. |
| **Ontologies** | A formal specification of a shared conceptualization, usually of domain-specific concepts. |
| **Part-of-Speech tagging** | Assigning of an appropriate morpho-syntactic class to a word. |
| **Precision (IR measure)** | Ratio of a number of relevant documents retrieved in the total documents retrieved. |
| **Pragmatics** | Study of how the language is used to give and receive information, and how it is used to express and understand author's intentions to one another. |
| **Query** | User's information need expressed in a specified syntax. |
| **Recall (IR measure)** | Number of relevant documents retrieved in the total number of relevant documents. |
| **Regular expression** | general pattern that allows to express alternative strings, repetitions and concatenations of substrings |
| **Relation finding** | Determination of relations between words, phrases and clauses (such as subject, object, location, etc.) |
| **Scatter/gather** | A browsing strategy which clusters the local documents in the answer set dynamically into topically-coherent groups and presents the user with descriptions of such groups. |
| **Semantics** | A study of how language organizes and expresses meaning. |
| **Shallow parsing** | Set of methods assigning a partial syntactic structure to a sentence, including typically P-o-S tagging, chunking and relation finding |
| **Small word technique** | A language identification technique based on the probability of an occurrence of frequent short words (typically not longer than 5 characters) in a given language. Probabilities are based on a corpus analysis including more than 1000 words common for each language. |
| **Stemming** | A simple technique for reducing words to their grammatical roots. |
| **Syntagms** | Ordered combination of P-o-S's that corresponds to a valid sentence. |
| **TF.IDF** | Term frequency / Inversed document frequency. In the information retrieval model, the weight of word expressing its for a document where it occurs is proportional to the frequency of occurrence in the document related to its total frequency in all documents. |
| **Vector Space Model** | A classic model of document retrieval based on representing documents and queries as vectors of index terms. The model adopts as foundation the notion of term-to-term independence. |

# 7. Index

# 8. References

[ACL]        http://www1.cs.columbia.edu/~acl/home.html

[aenalyst] Victor Lavrenko and Matt Schmill and Dawn Lawrie and Paul Ogilvie and David Jensen and James Allan. Mining of Concurrent Text and Time Series.

[Aggarval99]    C.C. Aggarval. Fast Algorithms for Projected Clustering. ACM SIGMOD, Philadelphia, 61-72. 1999.

[Agrawal96]    R. Agrawal. Fast Discovery of Association Rules. In: U.M.Fayyad et al. (Eds.) Advances in knowledge discovery and data mining, 1996.

[Ait-Mokhtar97]  S. Aït-Mokhtar and J.-P. Chanod. Incremental Finite-State Parsing. In: Proceedings of ANLP'97, Washington, pp.72-79, 1997.

[Allan98]   J. Allan, J. Carbonell, G. Doddington, J. Yamron, and Y. Yang. Topic detection and tracking pilot study: Final report. In Proceedings of the DARPA Broadcast News Transcription and Understanding Workshop, 1998.

[Anderson58]    W T Anderson. An Introduction to Multivariate Statistical Analysis. Wiley, New York, 1958.

[Ando01]   R. Ando. The Document Representation Problem: An analysis of LSI and Iterative residual rescaling. Thesis. Cornell Univ. 2001.

[Appelt93]     D.E. Appelt et al. FASTUS: A Finite-State Processor for Information Extraction from Real-World Texts. In: Proc. of IJCAI'93, Chambery, France, 1993.

[Au00]     Peter Au and Matthew Carey and Shalini Sewraz and Yike Guo and Stefan M. Rüger. New paradigms in information visualization. In: Research and Development in Information Retrieval, 2000.

[Baeza-Yates99] Baeza-Yates and Ribeiro-Neto. Modern Information Retrieval. 1999.

[Baldwin98]     J. F. Baldwin, T. P. Martin, and J. M. Rossiter. Time series modelling and prediction using fuzzy trend information. In Proceedings of Iizuka 1998.

[Baldwin00]     N. Baldwin et al. An Evaluation Roadmap for Summarization Research.

[Bergstrom00]    A. Bergstrom et al. Enhancing Information Retrieval by Automatic Acquisition of textual Relations using Genetic Programming. Proc. of the 5th Intl Conference on Intelligent User Interfaces. 2000.

[Berry97]  M.J.A Berry and G. Linoff. Data Mining Techniques: For Marketing, Sales and Customer Support. J. Willey & Sons, New York. 1997.

[Berson99]     A. Berson, S. Smith, and K. Thearling. Building Data Mining Applications for CRM. McGraw Hill. 1999.

[Berthold99]     M. Berthold and D.J. Hand. Intelligent Data Analysis: An Introduction. ISBN: 3540658084. MIT Press. 1999.

[Besancon01]    R. Besancon et al. Improving Text Representations through Probabilistic Integration of Synonymy Relations. In: Proceedings of the Intl. Symposium on Applied Stochastic Models and Data Analysis (ASMDA'2001), 1, pp. 200-205, 2001.

[Besancon02] R. Besancon and M. Rajman. Evaluation of a Vector Space Similarity Measure in Multi-Lingual Framework. In: 3rd Intl. Conference on Language Resources and Evaluation. Pages 1537-1542. 2002.

[Besancon02a] R. Besancon. Integration of syntactic and semantic knowledge in vector-based text representations, PhD Thesis, 2002.

[Blache-proof]   Philippe Blache (1998b) Proof Nets for Controlling Ambiguity. In: Natural Language Processing, in proceedings of ICTAI'98.1998.

[Brill95transformationbased]     Eric Brill. Transformation-Based Error-Driven Learning and Natural Language Processing: A Case Study in Part-of-Speech Tagging. In: Computational Linguistics vol. 21 n# 4. Pages 543-565. 1995

[Calude99]    C.Calude, K. Salomaa and S.Yu. Metric Lexical Analysis. Technical Report, 1999.

[Calzolari02towards]    Nicoletta Calzolari, Antonio Zampolli, Alessandro Lenci. Towards a Standard for a Multilingual Lexical Entry: The EAGLES/ISLE Initiative. In: Proceedings of the Third International Conference on Computational Linguistics and Intelligent Text Processing. Pages: 264-279. 2002.

[Carreras02] X.Carreras et al. Named entity extraction using AdaBoost. In: Proc. Of CoNLL'02, Taipei, Taiwan, 2002.

[Carbonell97] J.G. Carbonell, et al. Translingual Information Retrieval: A comparative Evaluation. In: Intl. Joint Conference on Artificial Intelligence IJCAI'97. Pages 708-715. 1997.

[Carlson01]    L. Carlson et. al. An Empirical Study of the Relation Between Abstracts, Extracts and the Discourse Structure of Texts. DUC2001 Meeting on Summary Evaluation, New Orleans, 2001.

[castell95filtering]    Nuria Castell and Angels Hernandez. Filtering Software Specifications Written in Natural Language. In: Portuguese Conference on Artificial Intelligence. Pages: 447-455. 1995.

[Chalmers93]    Matthew Chalmers. Using a Landscape Metaphor to Represent a Corpus of Documents. In: Proc. European Conference on Spatial Information Theory, 1993.

[Chanod96]    J.-P. Chanod and P. Tapanainen. A Robust Finite-State Parser for French. ESSLLI 1996.

[Charniak93]    E. Charniak. Statistical Language Learning. MIT Press, 1993.

[Chen02] H.H. Chen. Multilingual Summarization and Question Answering. Workshop on Multilingual Summarization and Question Answering, COLING'02, Taipei, Taiwan, 2002.

[Chen02a]    C. Chen. Visualization Of Knowledge Structures. In: Handbook of Software Engineering and Knowledge Engineering, 2002.

[Chen99visualizing] Chaomei Chen and Leslie Carr. Visualizing the Evolution of a Subject Domain: A Case Study. In: IEEE Visualization '99. Pages:449-452. 1999.

[Chi00]    Ed H. Chi. A Taxonomy of Visualization Techniques Using the Data State Reference Model. 2000.

[Church95]    K.W.Church. One Term or Two? In: Proc of the 18th Annual Intl. ACM SIGIR Conference on Research and Development in Information Retrieval, 1995.

[Clifton01]C. Clifton et al. TopCat: Data Mining for topic Identification in a Text Corpus. In: Principles of Data  Mining and Knowledge Discovery. 2001.

[CoNLL]    http://cnts.uia.ac.be/conll2003/

[Cryssmann02] B. Cryssmann et al. An Integrated Architecture for Shallow and Deep Processing. In: Proc. of 40th Annual Meeting of the ACL, pages 441-448, Philadelphia, PA. 2002.

[Daelemans02] W. Daelemans and V. Hoste. Evaluation of Machine Learning Methods for Natural Language Processing Tasks. In: 3rd Intl. Conference on Language Resources and Evaluation. Pages 755-760. 2002.

[Daelemans99] W. Daelemans, S. Buckholtz and J. Veenstra. Memory-based Shallow Parsing. Proc CoNLL-EACL'99. 1999.

[Daum03]    M. Daum, K.A. Foth and W. Menzel. Constraint-based Integration of Deep and Shallow Parsing Techniques. In: Proc. of 10t Conference of the European Chapter of ACL, EACL'03, Budapest, Hungary. 2003.

[Dawson74]    J.L Dawson. Suffix Removal and Word Connation. In: ALLC Bulletin, 2(3): 33-46, 1974.

[Day97]    D.Day et al. Mixed Initiative Development of Language Processing Systems. In: Proc. of 5th conference on Applied Natural Language Processing, Washington D.C. 1997.

[Deerwester90] S. Deerwester et al. Indexing by Latent Semantic Analysis. J. Amer. Soc. Inform. Sci. 41, 6, 391-407, 1990.

[dehaspe97mining]     L. Dehaspe and L. De Raedt. Mining Association Rules in Multiple Relations. In: Proceedings of the 7th International Workshop on Inductive Logic Programming Vol. 1297. Pages: 125-132. 1997.

[Diab02Unsupervised] Mona Diab and Philip Resnik. An Unsupervised Method for Word Sense Tagging using Parallel Corpora. In: 40th Anniversary Meeting of the Association for Computational Linguistics (ACL-02). 2002.

[Dempster77]     A. Dempster et al. Maximum likelihood from incomplete data via EM algorithm. Journal of the Royal Statistical Society, Series B, 39(1):1-38, 1977.

[DUC]     http://www-nlpir.nist.gov/projects/duc

[Evans96nounphrase]     David A. Evans and Chengxiang Zhai. Noun-Phrase Analysis in Unrestricted Text for Information Retrieval. In: Proceedings of the {ACL}-96, 34th Annual Meeting of the Association for Computational Linguistics. Pages: 17-24. 1996.

[Even-zohar00classification]     Y. Even-Zohar and D. Roth. A classification approach to word prediction. In: NAACL-2000, The 1st North American Conference on Computational Linguistics.Pages: 124--131. 2000.

[Faure98acquisition]     David Faure and Claire Nédellec and Céline Rouveirol. Acquisition of Semantic Knowledge using Machine learning methods: The System "ASIUM". 1998.

[Fayyad95]     U.M. Fayyad and R. Uthurusamy, Editors. Proc. of the 1st Intl Conference  on Knowledge Discovery in Databases and Data Mining. Menlo Park, California, 1995.

[Feldman95]     R. Feldman and I. Dagan. Knowledge Discovery in Textual Databases. In:  Proc. of the First Intl. Conference on Knowledge Discovery (KDD-95). ACM, Montreal, 1995.

[Feldman96]     R. Feldman and Haym Hirsh. Mining Associations in Text in the Presence of Background Knowledge. In: Proc of the 2nd Intl. Conference on Knowledge Discovery and Data Mining, 1996.

[Feldman98]     R. Feldman et al. Knowledge Management: A Text Mining Approach. In: Proc. of the 2nd. Intl. Conf. On Practical Aspects of Knowledge Management, PAKM'98, 1998.

[Fellbaum98towards]     Christiane Fellbaum. Towards a Representation of Idioms in WordNet. In: Use of WordNet in Natural Language Processing Systems: Proceedings of the Conference. Pages: 52-57. 1998.

[Fensel01oil]     D. Fensel and F. van Harmelen and I. Horrocks and D. McGuinness and P. Patel-Schneider. OIL: An ontology infrastructure for the semantic web. In: IEEE Intelligent Systems, 16(2):38--44. 2001.

[Fiore01treemap]     Fiore, Andrew and Marc Smith. "Tree Map Visualizations of Newsgroups", 2001.

[Frank03] A. Frank. Integrated Shallow and Deep Parsing: TopP meets HPSG. In: Proceedings of ACL'03, pp. 104-111, Sapora, Japan, 2003.

[Frawley91]     W.J. Frawley, G. Piatetsky-Shapiro and C.J. Matheus. Knowledge Discovery in Databases: An Overview. In: G. Piatetsky-Shapiro and W. J. Frawley (Eds.): Knowledge Discovery in databases, pages 1-27, MIT Press, 1991.

[Gaizauskas98] R. Gaizauskas and Y. Wilks. Information Extraction: Beyond Document Retrieval. Journal of Documentation, 54(1), 1998.

[Gaussier98]     E. Gaussier, G. Grefenstette and D.A.Hull. Xerox TREC-6 Site Report: Cross Language Text Retrieval. 1998.

[Goldstein99]     J. Goldstein. Summarizing Text Documents: Sentence Selection and Evaluation Metrics. In: ACM-SIGIR'99. Berkeley, CA, USA.1999.

[Grefenstette95] G. Grefenstette. Comparing two language identification schemes. In: Proc. of the 3rd Intl. Conference on the Statistical Analysis of the Textual Data JADT'95, Rome, Italy, 1995.

[Grefenstette98] G. Grefenstette. Cross-language Information retrieval. Kluwer Academic Publishers, 1998.

[Guha98] Guha et al. CURE: An Efficient Clustering Algorithm for Large Databases. ACM SIGMOD 1998

[Guha99] Guha et al. ROCK: A Robust Clustering Algorithm for Categorical Attributes. IEEE Conference on Data Engineering. 1999.

[Hahn99] U. Hahn and U. Heimer. Knowledge-based Text Summarization. In: I. Mani. And T. Maybury. Advances in Automated Summarization, MIT Press, 1999.

[Hahn00] U. Hahn and I. Mani. The Challenges of Automatic Summarization. IEEE Computer 33(11): 29-36, 2000.

[Habert98extending]     Habert B, Nazarenko A, Zweigenbaum P, and Bouaud J. Extending an existing specialized semantic lexicon. In: Rubio A, Gallardo N, Castro R, and Tejada A, eds, First International Conference on Language Resources and Evaluation, Granada. 1998:663--8. Also available at url http://www.biomath.jussieu.fr/~pz/biblio-pierre/#Habert:LREC98. 1998.

[Halkidi01] M. Halkidi, Y. Batistakis and M. Vazirgiannis. On Clustering Validation Techniques. In Journal of Intelligent Information Systems. 17(2-3):107-145, 2001.

[Han01]     J. Han and M. Kamber. Data Mining. Morgan Kaufmann Publishers. 2001.

[Hand01]     D.J. Hand, H. Mannila, P. Smyth. Principles of Data Mining: Adaptive Computation and Machine Learning. ISBN: 026208290X. MIT Press, 2001

[Hartigan75]     J. Hartigan. Clustering Algorithms. John Willey & Sons. New York, 1975.

[Hastie01]     T. Hastie, R. Tibshirani and J.H. Friedman. The Elements of Statistical Learning: Data Mining, Inference and Prediction. ISBN: 0387952845, Springer Verlag, 2001.

[Hastings99]     P.W. Hastings. How Latent is Latent Semantic Indexing. IJCAI'99. 1999.

[Havre]     Susan Havre and Beth Hetzler and Lucy Nowell. ThemeRiver(tm): In Search of Trends, Patterns, and Relationships.

[He00comparative]     Ji He and Ah-Hwee Tan and Chew-Lim Tan. A Comparative Study on Chinese Text Categorization Methods. In: PRICAI Workshop on Text and Web Mining. Pages:24-3. 2000.

[Hearst92]     M. Hearst and J. Pedersen. Acquisition of Hyponyms from Large Text Corpora. Proc. of 14th Intl. Conference on Computational Linguistics. 1992.

[Hearst95] Marti A. Hearst. TileBars: Visualization of Term Distribution Information in Full Text Information Access. In: Proceedings of the Conference on Human Factors in Computing Systems, CHI'95, 1995.

[Hearst96] Marti A. Hearst and Jan O. Pedersen. Reexamining the cluster hypothesis: Scatter/gather on retrieval results. In: Proceedings of SIGIR-96, 19th ACM International Conference on Research and Development in Information Retrieval, 1996

[Hearst99]     M. Hearst. Untangling Text Data Mining. In:Proc. of the 37th Annual Meeting of the Association ofComputational Linguistics, 3-10, 1999.

[Herman00]     Herman and G. Melançon and M. S. Marshall. Graph Visualization and Navigation in Information Visualization: A Survey. In:IEEE Transactions on Visualization and Computer Graphics, 2000.

[Hetzler98]     Beth Hetzler and Paul Whitney and Lou Martucci and Jim Thomas. Multi-faceted Insight Through Interoperable Visual Information Analysis Paradigms. In: Proceedings {IEEE} Symposium on Information Visualization 1998

[Hetzler98a]     B. Hetzler and W. Harris and S. Havre and P. Whitney. Visualizing the Full Spectrum of Document Relationships. In Proceedings of the Fifth    International Society for Knowledge Organization (ISKO) Conference, 1998.

[Hobbs93]     J. Hobbs. The Generic Information Extraction System. Proc. of the 5th Message Understanding Conference. Morgan Kaufman. 1993.

[Hoffmann99] T. Hoffmann. Probabilistic Latent Semantic Indexing. SIGIR'99. 50-57. 1999.

[Hotho01] A. Hotho, A. Maedche and S. Staab. Ontology-based Text Document Clustering. In: Proceedings of the IJCAI'01 Workshop "Text learning: beyond supervision", Seattle, 2001.

[Huber85] P.J. Huber. Projection Pursuit. The Annals of Statistics, Vol. 13, No.2 pages 435-474, 1985.

[Hull96method] David A. Hull and Jan O. Pedersen and Hinrich Schütze. Method combination for document filtering. In: Proceedings of SIGIR-96, 19th ACM International Conference on Research and Development in Information Retrieval. Pages:279-288. 1996.

[InfoCrystal] http://www.scils.rutgers.edu/~aspoerri/InfoCrystal/InfoCrystal.htm

[Jing99] H. Jing and E. Tzoukermann. Information retrieval based on context distance and morphology. SIGIR '99, Proceedings on the 22nd annual international ACM SIGIR conference on Research and development in information retrieval, pages 90-96, 1999.

[Joachims01] T. Joachims. A Statistical Learning Model of Text Classification with Support Vector Machines. Proceedings of the Conference on Research and Development in Information Retrieval (SIGIR), ACM, 2001.

[Joachims98] T. Joachims, Text Categorization with Support Vector Machines: Learning with many relevant features. In: European Conference on Machine Learning 1998.

[Joshi97treeadjoining] Joshi, A.K. and Schabes, Y.. Tree-Adjoining Grammars. In: Handbook of Formal Languages, G. Rozenberg and A. Salomaa (eds.), Vol. 3, Springer, Berlin, New York. 1997.

[Kaski97] S. Kaski. Data Exploration using Self-Organizing Maps. In Acta Polytechnica Scandinavica. 1997.

[Kaski98] S. Kaski. Dimensionality Reduction by Random Mapping: Fast Similarity Computation for Clustering. In: Proc. of IJCNN'98, IEEE Intl. Joint Conference on Neural Networks. 1998.

[Kaufman90] L. Kaufman and P. Rousseeuw. Finding groups in data: An Introduction to Cluster Analysis. John Wiley and Sons, New York, 1990.

[Keim00] D.A. Keim. Designing Pixel-Oriented Visualization Techniques: Theory and Applications. In: IEEE Transactions on Visualisation and Computer Graphics. Vol 6, No. 1, 2000. http://fusion.cs.uni-magdeburg.de/pubs/TVCG00.pdf

[Keim02] D.A. Keim. Information Visualization and Visual Data Mining. In: IEEE Transactions on Visualization and Computer Graphics, 7/1, 100-107, 2002.

[Kilgarriff-dont] A. Kilgarriff. I don't believe in word senses. In: Computers and the Humanities, forthcoming. 1997

[Klavans97natural] J. Klavans, C. Jacquemin, and E. Tzoukermann. A natural language approach to multi-word term conflation. In: Proceedings of the third Delos Workshop -- Cross-Language Information Retrieval. 1997.

[Kohonen95] T. Kohonen. Self-organizing Maps. Springer, Berlin 1995.

[Kool00] A.Kool et al. Genetic Algorithms for Feature Relevance Assignment in memory-based language processing. In: Proc. of 4th Conference on Computational Natural Language Learning and of the Second Learning Language in Logic Workshop, pages 103–106, Lisbon, 2000.

[Kraaij02] W. Kraaij, M. Spitters and A. Hulth. Headline extraction based on a combination of uni- and multidocument summarization techniques. DUC2002 Meeting on Text Summarization, Philadelphia, USA. 2002.

[Kraft03Textual] Donald H. Kraft, Jianhua Chen, Maria J. Martin-Bautista, Maria-Amparo Vila. Textual information retrieval with user profiles using fuzzy clustering and inferencing. In: Intelligent exploration of the web. Pages: 152 – 165. 2003.

[Krovetz93] R. Krovetz. Viewing Morphology as an Inference Process. In: R. Korfhage et al.: Proc. 16th ACM SIGIR Conference, pp. 191-202, Pittsburgh, June 27-July 1, 1993.

[Kukich92] K. Kukich. Techniques for Automatically Correcting Words in Text. In: ACM Computing Surveys, 24(4):377-439, 1992.

[Kurtz96] S. Kurtz. Approximate String Searching under Weighted Edit Distance. In Proceedings of Third South American Workshop on String Processing, Recife, Brazil, pages 156-170. Carlton University Press, 1996.

[Lassila99] O. Lassila and R. Swick, eds. Resource Description Framework (RDF) Model and Syntax Specification, W3C Recommendation, http://www.w3.org/TR/REC-rdf-syntax . 1999.

[Lebart98] L. Lebart et al. Exploring Textual Data. Kluwer Academic Publishers, ISBN 0-7923-4840-0, 1998.

[Lent97discovering] Brian Lent and Rakesh Agrawal and Ramakrishnan Srikant. Discovering Trends in Text Databases. In: Proc. 3rd Int. Conf. Knowledge Discovery and Data Mining, KDD. Pages: 227-230. 1997.

[Lewis96] D.D. Lewis et al. Training Algorithms for Linear Text Classifiers. In: Proc. of the 19th Annual Intl. ACM SIGIR Conference on Research and Development in Information Retrieval SIGIR'96. Pages 298-306. 1996.

[loper00applying] E. Loper. Applying semantic relation extraction to information retrieval. Master's thesis, Massachusetts Institute of Technology. 56. 2000

[Lovins68] J.B. Lovins. Development of a Stemming Algorithm. In : Mechanical Translation and Computational Linguistics, 11, 22-31, 1968.

[Lyberworld] http://www.darmstadt.gmd.de/~hemmje/Activities/Lyberworld/

[Mani01] I. Mani. Automatic Summarization. ISBN: 1588110591, John Benjamins Pub Co., 2001.

[Mani01a] I. Mani. Summarization Evaluation: An Overview. In.: Workshop on Automatic Summarization, NAACL, Pittsburg, 2001.

[Manning99] C.D.Manning and H. Schultze. Foundations of Statistical Natural Language Processing. MIT Press, 1999.

[Maulik02] U. Maulik and S. Bandyopadhyay. Performance Evaluation of Some Clustering Algorithms and Validity Indices. In : IEEE Transactions on Pattern Analysis and Machine Intelligence. 24(12):1650-1654. 2002.

[Mikheev00] Andrei Mikheev. Document Centered Approach to Text Normalization. In: Proc. of SIGIR, 2000.

[Miller98] S. Miller et al. Algorithms that learn to extract information – BBN-the system as used for MUC-7. In: Proc. of MUC-7, 1998.

[Mitkov02] R. Mitkov. Anaphora resolution. Longman 2002.

[Mladenic98feature] Dunja Mladenic. Feature Subset Selection in Text-Learning. In: European Conference on Machine Learning. Pages: 95-100. 1998.

[Mladenic96Web] Mladenic, D. Personal WebWatcher: Implementation and Design, Technical Report IJS-DP-7472, October, 1996.

[Monz01] Christof Monz, Maarten de Rijke. Introduction to Information Retrieval. In: ESSLI'01, Helsinki, Finland, August 2001.

[Morhi96fst] Mehryar Mohri. On Some Applications of Finite-State Automata Theory to Natural Language Processing. Natural Language Engineering, 2:1-20, 1996.

[Ng94] Effective and efficient clustering methods for spatial data mining. In: Proc. of the 20th conference on VLDB, 144-155, Santiago, Chile, 1994.

[Nigam99text] Kamal Nigam and Andrew K. McCallum and Sebastian Thrun and Tom M. Mitchell. Text Classification from Labeled and Unlabeled Documents using EM. In: Machine Learning Vol. 39 #. 2/3. Pages: 103-134. 2000.

[Nigam99using] Kamal Nigam, John Lafferty, and Andrew McCallum. Using maximum entropy for text classification. In: IJCAI-99 Workshop on Machine Learning for Information Filtering. Pages 61-67. 1999.

[NIRVE]    http://www.itl.nist.gov/iaui/vvrg/cugini/uicd/nirve-home.html

[Noreault81]    Terry Noreault, Michael McGill, and Matthew B. Koll. A performance evaluation of similarity measures, document term weighting schemes and representations in a Boolean environment. In R. N. Oddy, S. E. Robertson, C. J. van Rijsbergen, and P. W. Williams, editors, Information Retrieval Research, pages 57--76. Butterworths, London, 1981.

[Nurkkala94parallel]    Tom Nurkkala and Vipin Kumar. A Parallel Parsing Algorithm for Natural Language using Tree Adjoining Grammar. 1994.

[Oard97]  D.W. Oard. The State-of-the-Art in Text Filtering. In: User Modeling and User Adaptation Techniques, 7(2):141-178, 1997.

[Oflazer96]    K. Oflazer. Error-tolerant Finite State Recognition with Applications to Morphological Analysis and Spelling Correction. Computational Linguistics, 22(1), 1996.

[Ogden99]    W. Ogden. Getting information from documents you cannot read: An interactive cross-language text retrieval and summarization system. In SIGIR/DL Workshop on Multilingual Information Discovery and Access, Aug. 1999.

[OpenDX] http://researchweb.watson.ibm.com/dx/

[Paice90]  C.D. Paice. Another Stemmer. In: SIGIR Forum 24(3): 56-61. 1990.

[Pazienza97]    M.T. Pazienza. Information Extraction: A multidisciplinary Approach to an Emerging Information Technology. Springer-Verlag, 1997

[Platt98]    J. Platt. Sequential Minimal Optimization. A fast Algorithm for training support vector machines. Technical Report. Microsoft Research, 1998.

[Poibeau03]    T. Poibeau. Extraction Automatique d'Information du texte brut au web semantique. ISBN: 2-7462-0610-2. Lavoisier, Paris. 2003.

[Porter80]  M.F. Porter. An Algorithm for Suffix Stripping. In: Program, 14, 130-137, 1980.

[Poutsma01]    A.Poutsma. Applying Monte-Carlo Techniques to Language Identification. In: CLIN'01, 2001.

[Pottenger01]    William M. Pottenger and Ting-hao Yang. Detecting Emerging Concepts in Textual Data Mining. Computational Information Retrieval, Michael Berry, Ed., SIAM, Philadelphia, PA, August 2001.

[Rahm00]  E. Rahm and H.H. Do. Data Cleaning: Problems and Current Approaches. IEEE Bulletin of the Technical Committee on Data Engineering. 23(4), 2000.

[Rajaraman01]  Rajaraman And Ah-Hwee. Topic Detection, Tracking and Trend Analysis Using Self-organizing Neural Networks. In: PAKDD'01, 2001.

[Rajman97]    M. Rajman and Besançon R.. Text Mining: Natural Language techniques and Text Mining applications. Proc. of the 7th IFIP 2.6 Working Conference on Database Semantics (DS-7), IFIP Proceedings serie, Leysin (Switzerland), oct, 1997.

[Rajman98]    Rajman, M. and Lebart. Similarités pour données textuelles. In: 4th International Conference on Statistical Analysis of Textual Data , JADT'98. 1998.

[Rajman99]    M. Rajman and R. Besancon. Stochastic Distributional Models for Textual Information Retrieval. In: 9th International Symposium on Applied Stochastic Models and Data Analysis (ASMDA-99), 80-85, Lisbon, Portugal, June 14-17, 1999.

[Ratnaparkhi97] A. Ratnaparkhi. A Simple Introduction to Maximum Entropy Models for Natural Language Processing. IRCS Report 97--08, 1998.

[Robertson94] S. Robertson et al. Okapi at TREC-3. In: Proceedings of the 3rd Text Retrieval Conference, pp. 109-126, 1994.

[Rosario00]    B. Rosario. Latent Semantic Indexing: An Overview. 2000.

[Roth02]  V. Roth et al. A resampling approach to cluster validation. In: Computational Statistics, 2002.

[Sammon69]    J.W. Sammon: A nonlinear mapping for data structure analysis. IEEE
              Transactions on Computers C18, 401-409, 1969.

[Saraee95]    M. H. Saraee and B. Theodoulidis. Knowledge discovery in temporal
              databases. In IEE Colloquium on Digest No. 1995/021(A), pages 1/1--1/4, 1995.

[Saraee]      Mohammad Saraee and George Koundourakis and Babis Theodoulidis. Pattern
              Discovery In Time-Oriented Data.

[Schulz02] K.U. Schulz and S. Mihov: Fast String Correction with Levenshtein-Automata. In: Intl.
              Journal of Document Analysis and Recognition (IJDAR) 5(1):67-85, 2002.

[Sebastiani02] F. Sebastiani. Machine learning in automated text categorization. In: ACM
              Computing Surveys, 34(1):1-47, 2002.

[Sebrechts99]    Marc M. Sebrechts and John Cugini and Sharon J. Laskowski and Joanna
              Vasilakis and Michael S. Miller. Visualization of Search Results: A Comparative
              Evaluation of Text, 2D, and 3D Interfaces. In: Research and Development in
              Information Retrieval, 1999.

[SENSEVAL]        http://www.itri.brighton.ac.uk/events/senseval/

[Shang02] Yi Shang et al.: Precision Evaluation of Search Engines. In: WWW: Internet and Web
              Information Systems, 5, 159-173, 2002.

[Shivakumar00] Shivakumar Vaithyanathan Jianchang Mao Byron Dom. Hierarchical Bayes for
              Text Classification. PRICAI Workshop on Text and Web Mining. 2000.

[Schultz96lvcsrbased]    T. Schultz and I. Rogina and A. Waibel. LVCSR-based Language
              Identification. In: Proc. ICASSP '96. Pages: 781-784. 1996.

[Spark-Jones99] K. Sparck-Jones. Automatic Summarizing: Factors and Directions. In: I. Mani
              and M.T. Maybury, (Eds.), Advances in Automated Text Summarization. 1999.

[Smyth96]        P. Smyth. Clustering Using Monte Carlo Cross-Validation. Proc. of the KDD
              Conference 1996.

[Strzalkowski99] T. Strzalkowski et al. Evaluating Natural Language Processing Techniques in
              Information Retrieval. In: Strzalkowski (Ed.). Natural Language Information Retrieval,
              Kluwer Ac. Publishers, 1999.

[Sullivan01]        D. Sullivan. Document Warehousing and Text Mining. ISBN: 0-471-39959-0. J.W.
              Wiley, 2001.

[Takeuchi02use] Koichi Takeuchi and Nigel Collier. Use of Support Vector Machines in Extended
              Named Entity Recognition. In: 6th Conference on Natural Language Learning 2002
              (CoNLL-2002), pages 119—125. 2002.

[TREC]        http://trec.nist.gov

[Toivonen95pruning]        H. Toivonen, M. Klemettinen, P. Ronkainen, K. Hatonen, and H.
              Mannila.    Pruning and grouping of discovered association rules. In: Workshop Notes
              of the ECML-95 Workshop on Statistics, Machine Learning, and Knowledge Discovery
              in Databases, pp. 47-52, Heraklion, Greece. 1995.

[Uchimoto97]    K. Uchimoto, H. Ozaku & H. Isahara: "A Method for Identifying Topic-changing
              Articles in Discussiontype Newsgroups within the Intelligent Network News Reader
              HISHO" the Natural Language Processing Pacific Rim Symposium, NLPRS, 1997.

[Uszkoreit02]    H. Uszkoreit. New chances for deep linguistic processing. In: Proc of
              COLLING'02. 2002.

[Vapnik95]    V. Vapnik. The Nature of Statistical Learning Theory. Springer, New York, 1995.

[Veronis98]    J. Veronis and N. Ide. Word Sense Disambiguation: the State-of-the-Art. In: CL
              24(1):1-40. 1998.

[Rundensteiner02] E. A. Rundensteiner et al. XmdvTool: Visual Interactive Data Exploration and
              Trend Discovery of High-dimensional Data Sets. ACM SIGMOD 2002, Wisconsin, 2002.

[Sowa00Ontology]        John F. Sowa. http://users.bestweb.net/~sowa/ontology/

[VisIt]        http://www.visit.uiuc.edu/

[Voorhees99natural]    Ellen M. Voorhees. Natural Language Processing and Information Retrieval. In: SCIE. Pages: 32-48. 1999.

[Weiss98]  S. M. Weiss and N. Indurkhya. Predictive Data Mining: A practical Guide. ISBN 1-55860-403-0. Morgan-Kaufman. 1998.

[White97]  White, H. D., McCain, K. W. Visualization of literatures. In: Annual Review of Information Science Technology 32: 99-168, 1997

[Williams00]    C.K.I Williams. A MCMC Approach to hierarchical mixture modeling. Proc. of Neural Info Processing Systems (NIPS 2000). December 2000.

[Wong00] Pak Chung Wong and Wendy Cowley and Harlan Foote and Elizabeth Jurrus and Jim Thomas. Visualizing Sequential Patterns for Text Mining. 2000.

[Wong99visualizing]    Pak Chung Wong and Paul Whitney and Jim Thomas. Visualizing Association Rules for Text Mining. In: INFOVIS. Pages: 120-123. 1999.

[Yang94]  Y. Yang. An example-based mapping method for text-categorization and retrieval. ACM transaction on Information Systems, 12(3):252-277. 1994.

[Yang98]  Y. Yang et al. Translingual Information Retrieval: Learning from bilingual corpora. Artificial Intelligence. 103(1-2):323-345. 1998.

[Yang99a]    Y. Yand and Xin Liu. Re-examination of text classification methods. 1999

[Yang99b]    Y. Yang. An evaluation of statistical approaches to text categorization. In: Journal of Information Retrieval, Vol 1, No. 1/2, pp 67-88, 1999.

[Zhang96]    Zhang et al. BIRCH: An Efficient Method for Very Large Databases. ACM SIGMOD 1996.

# 9. Annex: NEMIS WG1 Progress Report

ÉCOLE POLYTECHNIQUE
FÉDÉRALE DE LAUSANNE

Project:        **NEMIS - Network of Excellence in Text Mining
and its Application in Statistics**

Document:  **NEMIS WG1 : Progress Report**

Date:        03 October 2003

Version:     0.1

Description: This document contains a detailed description of tasks
performed for the creation of the WG1 community.

**LIA-EPFL Artificial Intelligence Laboratory**
Institure of Core Computing Science
INR - CH-1015 Lausanne
Switzerland

# NEMIS WG1 : Progress Report

**Identification of key actors, products and commercial companies**

Identification of key actors in the field of Text Mining aims at identification of key personalities currently active in the field of text mining, of institutes where research is being conducted and applied, as well as of products developed either in the academic or commercial sphere. A part of this task has been dedicated to an active incorporation of key actors in the NEMIS project and creation of the WG1 community.

The following information sources have been exploited to obtain relevant raw data for processing:

- Recent workshops and conferences (1999 onwards) in TM and related areas (see Appendix A – List of Recent Workshops and Conferences)
- Directories maintained by existing or previous European projects (ELSNET and EUROMAP/HOPE, GRACE, NoE KDB) and other official directories (Francil, mlnet.org, universities)
- Conference papers, books and other published material, mainly from the citeseer.org source
- Internet subject catalogues (kdnuggets.com, google.com)
- Annual Reports of scientific organizations (CERN, LIA EPFL)
- Scientific Preprints databases (ArXiv.org, CERN Document Server, Cogprints)
- Scientific Journals
- ReferralWeb: Researcher clustering tool

The core data set contained individuals, products and companies currently active in TM research worldwide. The raw data gathering has focused on the recall maximization in order to address as much potential candidates as possible. [1] Upon this step, following criteria for selection have been set up, allowing to appropriately rank gathered items (criteria marked with an asterisk were used exclusively for ranking of authors):

| Criterion | Weight (0;1> |
| --- | --- |
| AllWebPage | 0.35 |
| ACI[2] | 0.15* |
| Recently published papers | 0.15* |
| Activity in Conferences | 0.15* |
| Academic References (ReferralWeb[3]) | 0.15* |
| Additional refinement | 0.05 |

Values for the "Autonomous Citation Indexing" and "Recently published papers" criteria were primarily extracted from the NECI Scientific Literature Digital Library Research Index (CiteSeer). "Activity in conferences" criterion relates specifically to conferences and workshops listed in appendix A, where authors of relevant papers and members of organizing committees were included. ReferralWeb has been selected as a relevant resource of academic reference being focused on scientists in the field of: artificial intelligence theory, natural language processing and information retrieval.

---

[1] Information gathering from identified resources and information extraction for relevant criteria was done using a wrapper software developed at LIA-EPFL, Lausanne and using a CDSware harvesting module developed at CERN, Geneva, Switzerland. http://cdsware.cern.ch/

[2] CiteSeer, http://www.neci.nec.com/~lawrence/aci.html

[3] Number of co-authors, http://foraker.research.att.com/refweb/version2/RefWeb.html

Technically, the evaluation of items was done using the MCDM (Multiple Criteria Decision Making) method [1] resulting in normalized weighted value over all criteria. For j-th person this value (NWV) then equaled to:

$$NWV_j = \sum_i [ [ ( v_{i,j} - min( v_i ) ) / ( max( v_i ) - min( v_i ) ) ] * w_i ] ,$$

where :

|  |  |
|---|---|
| $v_{i,j}$ | is a value of i-th criterion of j-th instance, |
| $max(v_i)$ and $min(v_i)$ | are border values for criterion i and |
| $w_i$ | is a weight of i-th criterion |

so the value has been normalized regarding all criteria included. As the sum of weights does not equal 1 for the list of products and commercial companies, further normalization was done in order to unify the value interval to <0;1> :

$$NWV_i' = NWV_i / ( \sum w_i )$$

This step of data sorting and ranking was focused on precision maximization in order to identify the "key actors" in the field that were consequently invited to participate at the 1st NEMIS Workshop (Launch Conference) that took place on the 5th of April 2003 in Patras, Greece. A lists of actors, commercial companies and products have been enclosed as a support material at this workshop, as well as a document containing an overview of topics in document processing and visualization in the domain of text mining.

**1st NEMIS Workshop (Launch Conference)**

In the WG1 group, the EPFL contribution has essentially concentrated on the production of an overview study describing the topics and techniques in the field of document processing and visualization. Before the first release of this document, an internal evaluation of its content has been carried out, based on a constructive interaction with the partners and members of the project. Corrections and completions have been made to deliver a document as exhaustive and understandable as possible.

However, additional efforts were still necessary to increase the exhaustivity and representativeness of the study. An additional bibliographic search based on the word wide web and other bibliographic sources has been carried out. This work presented interesting results and allowed us to discover missing topics in the document and achieve a better description of the document processing and visualization field. An additional study has resulted in description of visualization tools and techniques and the adjunction of a missing section on chronological data analysis that has been identified as an additional important task in the field of document processing, in particular to keep on the enrichment of the research topic descriptions and to enhance the coverage of the area.

As far as representativeness is concerned, it has been decided to submit the preliminary results of the study to the key actors in the field and to gather feedback. For this purpose a list of key actors has been compiled and an extensive mailing of the document is currently under preparation.

In addition, to improve the understandability of the different techniques described in the document, a search for concrete applications has been carried out and resulted in the inclusion of several application examples and graphical illustrations in the document. Finally, to improve readability and allow a fast navigation through the document, an exhaustive index and glossary have been added.

As for the event of Launch Conference, a respective presentation has been prepared, presenting the current state-of-the-art in document processing and visualization in the domain of text mining.

**Text Mining Roadmap**

The research topics in the domain of text mining as presented at the Launch Conference were taken as a source for a preparation of a detailed roadmap for follow-up research and technological development in the domain. The roadmap creation has been then essentially based on the analysis of the associated state-of-the-art, identification and refinement of relevant sub-domains in document processing and visualization.

Document processing and visualization were defined as multi-disciplinary tasks. Identification of their respective sub-domains has been based on the study of literature relevant for text mining and related disciplines, namely Information Retrieval (IR), Machine Learning (ML), Case-Based Reasoning (CBR), Statistical analysis, Knowledge Management (KM) and Natural language Processing (NLP). Identified sub-domains were roughly categorized as follows:

1. Document retrieval and selection
2. Document pre-processing and metadata production
3. Document warehousing
4. Data analysis
5. Visualization

The analysis of the state-of-the-art of the identified domains supported by the feedback as gathered from discussions at the Launch Conference lead to identification of the following research topics:

- Language identification and multilingual processing
- Relevance metrics
- Data integration for textual data
- Feature selection
- Alternatives to vector space model and related issues in document representation, mainly the similarity computation
- Stemming and lemmatization
- P-o-S tagging
- Parsing techniques (notably shallow vs. deep parsing, incremental robust parsing)
- Anaphora resolution
- Word sense disambiguation and related semantic issues (e.g. ontological interoperability)
- Cluster validation
- Entity tagging
- Term and relation extraction
- Exploitation of multivariate statistical techniques
- Summarization and question answering
- Trend analysis and chronological data analysis

[1] STEUER, R.E.: Multiple criteria optimization: theory, computation and application. 1. ed. New York, Wiley 1986.

**Appendix A – List of Recent Workshops and Conferences**

| | |
|---|---|
| ACL | Association for Computational Linguistics (series) |
| CIKM | Conference on Information and Knowledge Management |
| CLEF | Cross-Languages Evaluation Forum |
| COLING-ACL | Association for Computational Linguistics |
| Data Mining | Intl. Conference on Data Mining Methods and Databases for Engineering, Finance and other Fields |
| EACL | European Chapter of the Association of Computational Linguistics (series) |
| ECML/PKDD | European conference on Machine Learning/Principles of Knowledge Discovery in Databases |
| EMNLP | Empirical Methods in Natural Language Processing |
| FSKD | Fuzzy Systems and Knowledge Discovery |
| ICDM | Intl. Conference on Data Mining |
| ICONIP | Intl. Conference on Neural Information Processing |
| IDA | Intl. Symposium on Intelligent Data Analysis |
| IJCAI | Intl. Joint Conference on Artificial Intelligence |
| KDD | Intl. Conference on Knowledge Discovery and Data Mining |
| MLDM | IAPR Intl. Conference on Machine Learning and Data Mining |
| NLDB | Intl. Conference on Applications of Natural Languages to Information Systems |
| NLIS | Intl. Workshop on Natural Language and Information Systems |
| ODBASE | Intl. Conference on Ontologies, Databases and Applications of Semantics |
| PAKDD | Pacific-Asia Conference on Knowledge Discovery and Data Mining |
| PRICAI | Pacific Rim Intl. Conference on Artificial Intelligence |
| SDM/SIAM | SIAM Intl. Conference on Data Mining |
| SEAL | Simulated Evolution and Learning (C) |
| SIGIR | Intl. Conference on Research and Development in Information Retrieval |
| TextDM | Workshop on Text Mining (held at IEEE ICDM 2001) |
| TREC | Text Retrieval Conference (2000, 2001) ; TREC 2002 not available |
| DUC | Workshop on Text Summarization (2001, 2002) |
| | Workshop on Visualization and Data Mining |

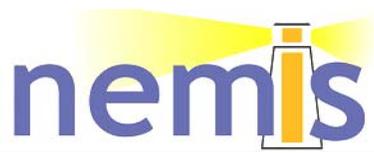